%% file: main.tex
\pdfoutput=1
\documentclass[11pt]{article}

\usepackage[]{naacl2021}

\usepackage{times}
\usepackage{latexsym}

\usepackage{graphicx}
\usepackage{booktabs}
\usepackage{enumitem}
\usepackage{xspace}
\usepackage{amsmath,amsthm,amsfonts,amssymb,bm,stmaryrd} 
\usepackage{multirow, makecell}
\usepackage{soul}
\usepackage{xcolor}
\usepackage{calc}
\usepackage[hang,flushmargin]{footmisc}
\usepackage{pifont}
\usepackage{subcaption}
\usepackage{tabularx} % used to handle word wrapping

\usepackage{tikz-qtree, tikz}
\usetikzlibrary{shapes,backgrounds,fit}
\usepackage[utf8]{inputenc} 
\usetikzlibrary{decorations.pathreplacing,arrows,shapes,positioning,shadows,calc}
\usetikzlibrary{decorations, decorations.text,backgrounds}
\tikzset{every picture/.style={font issue=\footnotesize},
    font issue/.style={execute at begin picture={#1\selectfont}}
}

\usepackage{pbsi}
\usepackage[T1]{fontenc}
\usepackage[utf8]{inputenc}

\usepackage{microtype}

\title{Get Your Vitamin C! \\ Robust Fact Verification with Contrastive Evidence}

\author{Tal Schuster \quad Adam Fisch \quad Regina Barzilay \\
  Computer Science and Artificial Intelligence Laboratory\\
  Massachusetts Institute of Technology \\
  {\tt \{tals,fisch,regina\}@csail.mit.edu} \\}

\input{macros}

\begin{document}
\maketitle

\input{sections/abstract}

\input{sections/introduction}
\input{sections/related}
\input{sections/data_collection}
\input{sections/tasks}

% \input{sections/models}
\input{sections/Experiments}

\input{sections/discussion}
\input{sections/acknolegments}
% \input{sections/ethical}

\bibliography{anthology,custom}
\bibliographystyle{acl_natbib}

% \clearpage
\appendix
% \onecolumn
\input{appendix/data_stats}

\input{appendix/setting}
\input{appendix/GPT}
\input{appendix/more_res}
\input{appendix/outputs}

% This is an appendix.

\end{document}

%% file: macros.tex
\newcommand{\tabref}[1]{Table \ref{#1}}
\newcommand{\figref}[1]{Figure \ref{#1}}

\newcommand{\secref}[1]{Section \ref{#1}}
\renewcommand{\paragraph}[1]{\vspace{0.15cm}\noindent\textbf{#1}}
\newcommand{\tpf}[1]{\noindent\textbf{#1}}

\DeclareMathOperator{\claims}{\mathcal{C}}
\DeclareMathOperator{\evs}{\mathcal{S}}

\newcommand{\revtup}{(s_{t-1}, s_t)}

\DeclareMathOperator{\rel}{rel}
\DeclareMathOperator{\SUP}{SUP}
\DeclareMathOperator{\REF}{REF}
\DeclareMathOperator{\NEI}{NEI}

\newcommand{\thedataset}{\textsc{VitaminC}\xspace}

\newcommand{\cmark}{\text{\ding{51}}}
\newcommand{\xmark}{\text{\ding{55}}}

\definecolor{darkgreen}{rgb}{0.0, 0.50, 0.50}

%% file: sections/abstract.tex
\begin{abstract}

Typical fact verification models use retrieved written evidence to verify claims. Evidence sources, however, often change over time as more information is gathered and revised. In order to adapt, models must be sensitive to subtle differences in supporting evidence. We present $\thedataset$, a benchmark infused with challenging cases that require fact verification models to discern and adjust to slight factual changes. We collect over 100,000 Wikipedia revisions that modify an underlying fact, and leverage these revisions, together with additional synthetically constructed ones, to create a total of over 400,000 claim-evidence pairs. Unlike previous resources, the examples in $\thedataset$ are \emph{contrastive}, i.e., they contain evidence pairs that are nearly identical in language and content, with the exception that one supports a given claim while the other does not. We show that training using this design increases  robustness---improving accuracy by 10\% on adversarial fact verification and 6\% on adversarial natural language inference (NLI). Moreover, the structure of \thedataset leads us to define additional tasks for fact-checking resources: tagging relevant words in the evidence for verifying the claim, identifying factual revisions, and providing automatic edits via factually consistent text generation.\footnote{The \thedataset dataset and our models are available at: \url{https://github.com/TalSchuster/VitaminC}}

\end{abstract}

%% file: sections/introduction.tex
\section{Introduction} \label{sec:intro}

    Determining the truthfulness of factual claims by comparing them to textual sources of evidence has received intense research interest in recent years. 
    An underlying, but often overlooked, challenge for this paradigm, however,
    is the dynamic nature of today's written resources. An extraordinary amount of new information becomes available daily; as a result, many consequential facts are established, changed, or added to over time. 
    We argue that the quality of fact verification systems should be measured by how well they adjust to new evidence. In this way, we seek to advance fact verification by requiring that models remain reliable and robust to the change present in practical settings.

% An underlying challenge for this paradigm is the time-sensitive quality of today's written resources. An extraordinary amount of new information becomes available daily; as a result, many consequential facts are established, changed, or are added over time. This can be especially prevalent for current events of significant public interest (for which fact verification is critical). For example, during the week of November 3rd, 2020, the Wikipedia article on the 2020 US presidential election was updated more than 600 times.\footnote{\url{https://bit.ly/2020_elec_edit_stat}} Standard fact verification evaluation benchmarks, however, assume a fixed time-stamp of independent and identically distributed claim and evidence pairs. As a result, they fail to test whether small, but factual, changes in evidence result in predictions that change in accordance.Ensuring that fact verification systems are \emph{sensitive} to potentially subtle-yet-key changes is essential for robust and reliable performance.

\begin{figure}[t]
    \centering
    % \includegraphics[width=0.4\textwidth]{figures/intro.png}
    % \resizebox{0.5\textwidth}{!}{\input{figures/intro.tex}}
    \includegraphics[width=0.48\textwidth]{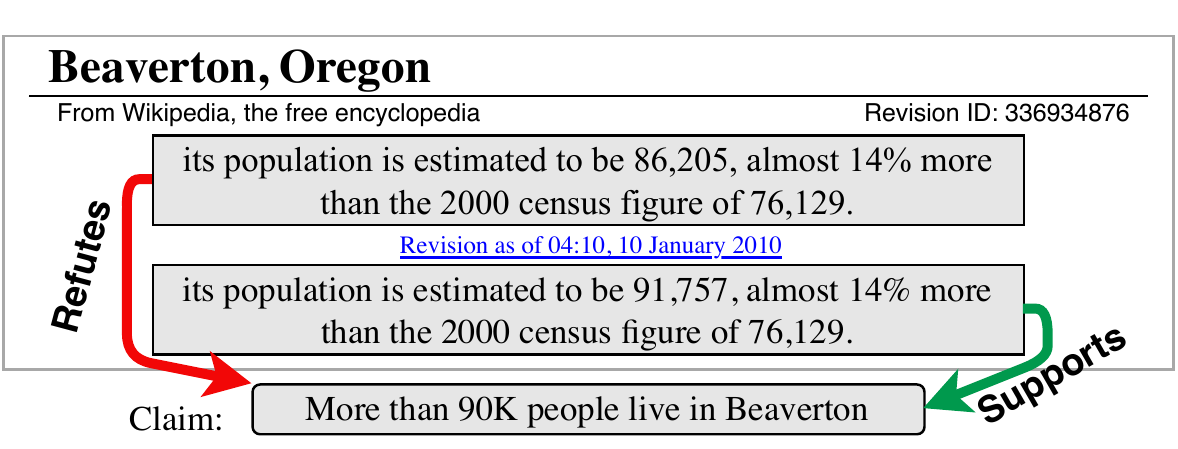}
    % \vspace{-1\baselineskip}
    \caption{In \thedataset, we focus on Wikipedia revisions in which the \emph{factual content} changes. This example revision now \emph{supports} an initially \emph{refuted} claim.}
    \label{fig:intro}
    \vspace{-1\baselineskip}
\end{figure}

To this end, we focus on fact verification with \emph{contrastive evidence}. That is, we infuse the standard fact verification paradigm with challenging cases that require models to be sensitive to factual changes in their presented evidence (hereon referred to interchangeably as ``context''). We present \thedataset,\footnote{Etymology of \thedataset: \emph{Contrastive} evidence keeps fact verification models robust and healthy, hence ``Vitamin C.''} a new large-scale fact verification dataset that is based on factual revisions to Wikipedia. The key concept is exemplified in Figure~\ref{fig:intro}: there a factual revision yields a contrastive pair of contexts that are nearly identical in language and content---except that one context refutes the given claim, while the other supports it.
% This novel benchmark requires successful verifiers to rely on external evidence during inference. 
%

This type of contrastive structure exposes existing deficiencies in model behavior. To illustrate this, we train a classifier on the popular FEVER fact verification dataset~\citep{thorne-etal-2018-fever} and evaluate it on contrastive claim-evidence pairs. We find that the model flips its prediction from the original verdict on only 56\% of the contrastive cases. When examples from \thedataset are included during training, however, the model's sensitivity increases, flipping on 86\% of contrastive cases. 

Such context-sensitive inference has two main benefits. First, it ensures that the model considers the provided evidence rather than relying on built-in static knowledge, such as that obtained via language model pre-training~\citep{petroni-etal-2019-language, roberts-etal-2020-much}. This is particularly important for scenarios in which the source of truth is mutable (e.g., the current US president, or new declarations as in \figref{fig:intro}). Second, this setting discourages certain biases and idiosyncrasies---such as exploiting differences in how true vs.\ false claims are posed---that are common in similar crowd-sourced datasets~\citep{poliak-etal-2018-hypothesis, schuster-etal-2019-towards}. 
Indeed, we show that augmenting both fact verification models and NLI models with \thedataset data improves their robustness to adversarial inputs.

Furthermore, our emphasis on contrastive contexts allows us to expand on the scope of commonly considered tasks.
Most of the fact verification literature focuses on resolving claims to be true or false~\citep{popat-etal-2018-declare, thorne-vlachos-2018-automated, wang-2017-liar}. 
The surrounding ecosystem, however, includes additional challenges, some of which we explore here: Documents such as Wikipedia articles are updated frequently; which edits represent factual changes? For a given claim and (refuting or supporting) evidence pair, which words or phrases in the evidence are most relevant? If we know that a certain claim is true, can we modify an out-dated document to be consistent with it?
We show that the unique structure of our \thedataset dataset can be leveraged to provide both supervised and distantly supervised data for these new questions.

%
% In the following, we describe the \thedataset dataset and its collection and curation process (\S\ref{sec:data_collection}),  define an expanded suite of tasks that fact verification with dynamic context engenders (\S\ref{sec:tasks}), and  present baseline models (\S\ref{sec:models}) with their results (\S\ref{sec:experiments}).
%
%

Our key contributions are as follows:\vspace{-5pt}
\begin{enumerate}[leftmargin=*, noitemsep]
    \item We pose a contrastive fact verification paradigm that requires sensitivity to changes in data; \vspace{3pt}
    \item We introduce \thedataset, a new large-scale dataset that supports this paradigm;\vspace{3pt}
    \item We demonstrate that training on \thedataset leads to better performance on standard tasks;\vspace{3pt}
    \item We show how \thedataset opens the door to additional research directions in fact verification.
\end{enumerate}

%% file: sections/related.tex
\section{Related Work} \label{sec:related}

\tpf{Fact Verification.}
The FEVER dataset~\citep{thorne-etal-2018-fever} fueled the development of many fact-checking models~\cite[e.g., see][\emph{inter alia}]{hanselowski-etal-2018-ukp, nie-etal-2019-simple, Nie_Chen_Bansal_2019, yoneda-etal-2018-ucl}. 
The claim creation process, however, required crowd-workers to write claims related to Wikipedia articles, and was found to engender biases that allow an evidence-agnostic model to achieve unexpectedly high performance~\citep{schuster-etal-2019-towards}. 
Other recent datasets cover verification against tables~\citep{Chen2020TabFact}, relational databases~\citep{aggchecker}, Wikipedia references~\citep{sathe-etal-2020-automated}, multiple articles~\citep{jiang-etal-2020-hover}, and search snippets~\citep{augenstein-etal-2019-multifc}. These resources all assume static ground truths.
% \citet{chen-etal-2019-seeing} collected subjective claims and opinions paired with references that provide different levels of agreement.
In contrast, $\thedataset$ compares objective claims to a dynamic source of truth, and requires models to change their verdicts accordingly.

% \paragraph{Adversarial and Bias-reduced Methods.}
\paragraph{Annotation Bias.}
Annotation artifacts are common in many NLP datasets, and affect performance on adversarial and contrastive examples~\citep{gardner-etal-2020-evaluating,ribeiro-etal-2020-beyond,ross2020explaining}.
Sentence-pair inference tasks such as fact verification~\citep{paul-panenghat-etal-2020-towards, schuster-etal-2019-towards} and NLI~\citep{gururangan-etal-2018-annotation, mccoy-etal-2019-right, poliak-etal-2018-hypothesis, tsuchiya-2018-performance} are no exception. Alleviating this bias requires either modeling solutions~\citep{karimi-mahabadi-etal-2020-end, pratapa-etal-2020-constrained, shah2020automatic, thorne2020avoiding, utama-etal-2020-towards}, which have limited effectiveness~\citep{utama-etal-2020-mind}, or adversarially removing troublesome training examples~\citep{bras2020adversarial} or manually collecting new ones~\citep{nie-etal-2020-adversarial, thorne-etal-2019-evaluating}, which is model specific. 
Instead, our dataset design avoids single-sentence artifacts and provides model-agnostic challenging examples that increase the robustness of trained models.

% rationales
\paragraph{Explainability.}
Current fact verification datasets provide sentence-level rationales~\citep{deyoung-etal-2020-eraser, petroni2020kilt} but do not enforce the model's verdict to rely on them---leading to a potential discrepancy. $\thedataset$ ensures the verdict is conditioned on the retrieved evidence.
%by making the model adjust its predictions accordingly. 
Moreover, we use the revision history as distant supervision for word-level rationales, allowing for finer-grained explanations~\citep{NEURIPS2018_4c7a167b, lei-etal-2016-rationalizing, portelli-etal-2020-distilling, thorne-etal-2019-generating}.

\paragraph{Factually Consistent Generation.}
Generating texts that match given facts is a known challenge~\citep{fan-etal-2020-generating, kryscinski-etal-2020-evaluating, lewis2020retrievalaugmented, parikh-etal-2020-totto, shah2020automatic, tian2020sticking} as language models tend to degenerate and hallucinate~\citep{Holtzman2020The, schuster-etal-2020-limitations, zhou2020detecting}. Moreover, evaluation is non-trivial, and usually manual. 
$\thedataset$ includes supervised data for training sequence-to-sequence models, and provides automatic evaluation via the fact verification classifier.

% Degeneration of language models is a known problem~\citep{Holtzman2020The}, potentially leading to hallucinations and factually false generations~\citep{schuster-etal-2020-limitations}.

% \citep{shah2020automatic}
% \citep{kryscinski-etal-2020-evaluating}
% \citep{fan-etal-2020-generating}
% \citep{lewis2020retrievalaugmented}
% \citep{parikh-etal-2020-totto}
% \citep{tian2020sticking}

% \tal{todo}

% fact verification

% biases and adversarial

%% file: sections/data_collection.tex
\input{tables/revision_examples}

\section{The \thedataset Dataset} \label{sec:data_collection}
% Our goal is to create a benchmark that requires successful fact verification classifiers to \emph{rely on external evidence} when predicting the verdict. Relying on \emph{context} rather than inherent dataset biases or built-in knowledge (e.g., via language model pre-training) is important for scenarios in which the source-of-truth is dynamic. For example, this can occur when verifying (1) facts that change over time (e.g., the current president of the US), or (2) facts that were updated due to new studies or observations (e.g., the effectiveness of certain medicines). 

 \thedataset (abbreviated {VitC}) is based on revisions to English Wikipedia. Wikipedia has become a comprehensive online resource that is rigorously maintained by a large and active community~\citep{Benjakob2019From}. 
 While adversaries do try to insert disinformation, popular pages are usually quickly corrected~\citep{wikihoax}.
 Furthermore, Wikipedia's policies dictate that its content should be written from a neutral perspective---or should otherwise objectively state all points of view.\footnote{\url{https://bit.ly/Wiki_Neutral_POV}} These properties make Wikipedia a suitable source of evidence for fact verification models. In the following section, we outline our process for mining factual revisions from Wikipedia.

% \begin{figure}[t]
%     \centering
%     \includegraphics[width=0.5\textwidth]{figures/collection_diag.pdf}
%     \caption{Diagram of the \thedataset data collection and annotation process.\tal{TODO: improve graphics, exact numbers, move to appendix?}}
%     \label{fig:collection_diag}
% \end{figure}

\subsection{Collecting Factual Revisions}
We collected the 5K most-viewed English Wikipedia articles\footnote{\url{https://bit.ly/Wiki_popular_pages}} as of January 2020, along with any additional articles referred from them (on average 100 per article). We also included all articles from the FEVER dataset~\cite{thorne-etal-2018-fever}. For each article, we retrieved up to 500 of its most recent revisions. In May 2020, we added all COVID-19 related articles\footnote{ \mbox{\url{https://wikimediafoundation.org/covid19}}} and all of their 41K revisions at the time. Combined together, this resulted in a total of $\sim$200 million revisions. For each revision, we identified all of the modified sentences and stored two versions: (1) before, and (2) after the edit.

In our task, we are only interested in edits made with an intent to introduce a \emph{factual} modification---i.e., a change for which one can make a claim that is supported by one sentence, but not by the other.\footnote{Many edits only reflect grammatical corrections, paraphrasing, or ``Wikification'' (text formatting/page linking).} To expedite annotation, we trained a BERT classifier~\citep{devlin-etal-2019-bert} on a small labeled set of revised sentences determined to be factual~\citep{yang-etal-2017-identifying-semantic}, and used this model to select the top 305K edited sentences from the corpus for manual annotation.
Trained human annotators were then presented with the sentence pairs, and were asked to mark the ones that indeed represented a factual change. Sentences lacking self-contained context were filtered (e.g., short expressions from tables or bulleted lists). 
%In total, $35\%$ of the revised sentences were flagged as factual updates, resulting in 107,056 sentence pairs that we use for the creation our fact verification dataset. 
Example annotations are presented in \tabref{tab:example_revisions}. Note that these annotations can also be recursively recycled for re-training the automated BERT classifier in the future to expand the corpus further  (we also introduce this as a task, see \S\ref{sec:flagging}).

\subsection{Writing Claims} \label{sec:write_claims}
The factual Wikipedia revisions guide us in creating challenging claims for fact verification. For each revision, annotators were asked to write two symmetric claims related to the same edit:\vspace{-3pt}
\begin{enumerate}[leftmargin=*, noitemsep]
    \item The first should be supported by the \emph{original} sentence and refuted by the \emph{revised} sentence;\vspace{2pt}
    \item The second should be supported by the \emph{revised} sentence and refuted by the \emph{original} sentence.\vspace{-5pt}
\end{enumerate}
When an explicit contradiction was not possible, a \emph{not enough information} (NEI) relation was used.
A group of 70 native English speakers\footnote{We sourced our annotators through TransPerfect.} wrote and reviewed claims. During the annotation period, annotations were delivered in weekly batches, from which we examined random samples to provide feedback and request corrections.
Annotators were instructed to write short and self-contained claims. Furthermore, annotators were instructed to avoid copying exact phrases and values when possible, in order to avoid a bias for substantially higher word overlap in supporting pairs over refuting pairs. For example, rather than stating, ``\emph{there are \underline{$x$} confirmed cases of coronavirus in the US}'', one can write ``\emph{there are \underline{more than $z$} confirmed cases of coronavirus in the US}'', which is supported if $x > z$ and refuted otherwise.
For revisions that only add new information or that remove outdated facts without replacing them, annotators wrote a single claim.

% Of course, a revision might simply add new information (see \figref{fig:intro}), or remove an outdated fact (without replacing it). For such cases, the annotators wrote a single claim.
% that is supported by one sentence, and marked the other sentence as  \textit{neutral}.
% Also, a few cases required a more careful annotation. For example, a Wiki sentence stating ``more than $x$'' supports a claim saying ``more than $y$'' only if $x>y$. When $x<y$, the relation is neutral since there is uncertainty whether the actual value is in the range of $[x,y]$ or $[y, \infty]$. 

\subsection{Adding Synthetic Revisions}
Naturally, the real Wikipedia revisions we collect mostly describe facts that frequently change over time, or that are prone to mistakes and corrections (such as quantitative values, see Appendix~\ref{sec:claim_cat}) \citep{faruqui-etal-2018-wikiatomicedits, yang-etal-2017-identifying-semantic}.
Sensitivity to contrastive contexts, however, is desirable behavior for any claim. This can both ensure consistency with external sources of truth, and improve the model's faithfulness via connecting the verdict with a specific evidence~\citep{jacovi-goldberg-2020-towards, ross2020explaining}.
For example, we require the model to not only classify the claim ``\emph{Tom Hanks was honored by a president}'' as true, but to also change its verdict to false if paired with a (fictional) contrasting evidence. As a result, we can verify that the model prioritizes sentence-pair inference over memorization, which can help it generalize better. % to other examples that might not fall within its training data.
% Such behavior can both ensure consistency with external sources-of-truth, and improve the model's explainability (connecting the verdict with the specific evidence).
% While these facts are perhaps the ones that require the most careful verification, across the spectrum of facts, we want our models to be sensitive to contrastive contexts for all claims.\footnote{For example, the claim ``\emph{the sky is blue}'' might be resolvable simply based on its status as a commonly known axiom. In certain contexts, however, this assertion \emph{can} be false (e.g., based on the time of the day or weather). Here, we want the model to modify its verdict by the context.} 
%\footnote{\url{https://en.wikipedia.org/wiki/Wikipedia:You_do_need_to_cite_that_the_sky_is_blue}}
Therefore, we use the FEVER dataset to augment \thedataset with \emph{synthetic} revisions to Wikipedia sentences.

% One limitation of Wikipedia revisions is that they do not cover facts that rarely change. \adam{There's a bit of a messaging mismatch here... do we mean to say that there are a certain class of facts that are dynamically changing that we should focus on? If it is a fact that is static (i.e., once established, will not change, like a law of nature), then why should we require models to use context for these? Why not rely on fixed internal LM KB?}. In order to increase the diversity of claims in our dataset and include facts with high interest but do not frequently change, we create \textit{synthetic} revisions. 

We follow the setting of \citet{schuster-etal-2019-towards} to expand claim-evidence pairs from FEVER~\citep{thorne-etal-2018-fever}. Specifically, given a false claim from FEVER, we ask annotators to edit the sentence that refutes it so that it will then support the originally false claim. Additionally, we ask them to write a new claim that is refuted by the new, modified sentence, but that is supported by the original version. Following this method, we obtain two claims where each can be supported or refuted by the original, or the synthetically revised, sentence. We follow the same process for constructing synthetic examples using true claims, but with flipped labels.

\subsection{Dataset Statistics}
In total, 304,671 revised Wikipedia sentences were examined by annotators, of which 107,056 (35\%) were found to express a factual modification and were passed to the group of expert annotators for claim writing. As two symmetric claims with opposing facts were created (when possible) for each revision, this resulted in a total of 325,724 total claim-evidence pairs. We collected 163,180 additional pairs following the synthetic process. The data was partitioned as shown in \tabref{tab:data_stats}. The assignment was done randomly by article, and is consistent with FEVER for overlapping articles. Appendix~\ref{app:stats} contains additional details.

\input{tables/statistics}

%% file: tables/revision_examples.tex
\setuldepth{Berlin}
\definecolor{bittersweet}{rgb}{1.0, 0.44, 0.37}
\definecolor{applegreen}{rgb}{0.55, 0.71, 0.0}

%Define a reference depth. 
%You can choose either relative or absolute.
%--------------------------
\newlength{\DepthReference}
\settodepth{\DepthReference}{a}%relative to a depth of a letter.
%\setlength{\DepthReference}{6pt}%absolute value.

% %Define a reference Height. 
% %You can choose either relative or absolute.
% %--------------------------
\newlength{\HeightReference}
\settoheight{\HeightReference}{a}
%\setlength{\HeightReference}{6pt}

%--------------------------
\newlength{\Width}%

\newcommand{\removebox}[2][bittersweet]%
{%
    \settowidth{\Width}{#2}%
    \colorbox{#1}%
    {%      
        \raisebox{-\DepthReference}%
        {%
                \parbox[b][\HeightReference+\DepthReference][c]{\Width}{\centering#2}%
        }%
    }%
}

\newcommand{\addbox}[2][applegreen]%
{%
    \settowidth{\Width}{#2}%
    \colorbox{#1}%
    {%      
        \raisebox{-\DepthReference}%
        {%
                \parbox[b][\HeightReference+\DepthReference][c]{\Width}{\centering#2}%
        }%
    }%
}

\newcommand{\edit}[1]{\textcolor{blue}{\ul{#1}}}

\begin{table*}[!t]
  \small
  \centering
  
%   \begin{tabular}{p{3cm}|p{11cm}}
%     \toprule
%       Factual & ...\\
%       & ...\\
%     %\bottomrule
%   \end{tabular}
  \begin{tabular}{p{0.8cm}|p{1cm}| p{13cm}}
    \toprule
     Factual & \multicolumn{2}{l}{Wikipedia sentences \emph{before} and \emph{after} a revision, presented with $\thedataset$ claims if the revision is factual.}  \\
    \midrule
      \multicolumn{1}{c|}{\xmark} & Before & More \edit{stringent} actions were taken in China once the \edit{seriousness} of the outbreak became apparent, such as quarantining entire cities affecting 60 million individuals in Hubei, and strict travel bans. \\
      \cmidrule{2-3} 
       & After &  More \edit{drastic} actions were taken in China once the \edit{severity} of the outbreak became apparent, such as quarantining entire cities affecting 60 million individuals in Hubei, and strict travel bans. \\

    \midrule
     \multicolumn{1}{c|}{\cmark} & Before & In animals, spaying involves an invasive removal of the ovaries, but rarely has major complications \edit{other than that spayed animals tend to gain weight}. \\
      \cmidrule{2-3} 
       & After &  In animals, spaying involves an invasive removal of the ovaries, but rarely has major complications; \edit{the superstition that it causes weight gain is not based on fact}.  \\
       \cmidrule{2-3} 
       & Claim 1 & Spayed animals gain weight. \\
       & Claim 2 & Weight gain in spayed animals is a superstitious myth. \\

    % \midrule
    %   Factual & $s_{t-1}$ & 432 Park Avenue is the \st{third}-tallest building in New York City and will be the tallest residential building in the Western Hemisphere once opened. \\
    %   \cmidrule{2-3} 
    %   & $s_{t}$ &  432 Park Avenue is the \ul{second}-tallest building in New York City and will be the tallest residential building in the Western Hemisphere once opened. \\
    %   \cmidrule{2-3} 
    %   & $c_1$ & 432 Park Avenue comes third among tall buildings in New York City. \\
    %   & $c_2$ & 432 Park Avenue comes second among tall buildings in New York City. \\
       
     \midrule      
          \multicolumn{1}{c|}{\xmark} & Before & As of 16 March, more than 182,000 cases of the disease have been reported in over  160 countries and territories, resulting in \edit{around 79,000 recoveries and more than 7,100 deaths}. \\
      \cmidrule{2-3} 
       & After &  As of 16 March, more than 182,000 cases of the disease have been reported in over  160 countries and territories, resulting in \edit{more than 7,100 deaths and around 79,000 recoveries}. \\
       
        \midrule  
          \multicolumn{1}{c|}{\cmark} & Before & Global hybrid sales are led by the Prius family, with sales of \edit{4.7} million units representing \edit{66.8\%} of TMC worldwide sales of \edit{7.053} million Lexus and Toyota units through \edit{September} 2014. \\
      \cmidrule{2-3} 
       & After &  Global hybrid sales are led by the Prius family, with sales of \edit{5.264} million units representing \edit{65.4\%} of TMC worldwide sales of \edit{8.048} million Lexus and Toyota units delivered through \edit{July} 2014. \\
       \cmidrule{2-3} 
       & Claim 1 & Prius sold less than 5 million units, representing over 65.5\% of TMC worldwide sales. \\
       & Claim 2 & Prius sold more than 5 million units, representing less than 65.5\% of TMC worldwide sales. \\

    \bottomrule

    \end{tabular}
  \caption{Examples of \emph{non-factual} revisions vs. \emph{factual} revisions, and the claims associated with the later. Factual updates change the outcome (i.e., true or false) of a claim that might be in question. Accordingly, the verdict of a classifier should change based on the version presented. Modified words are underlined and colored.}
  \vspace{-10pt}
  \label{tab:example_revisions}
\end{table*}

%% file: tables/statistics.tex
\begin{table}[t]
    \centering
    \resizebox{\columnwidth}{!}{%
    \begin{tabular}{l|rr|rr|rr}
    \toprule
        & \multicolumn{2}{c|}{Supports} &  \multicolumn{2}{c|}{Refutes} & \multicolumn{2}{c}{NEI} \\
        Split & Real & Syn & Real & Syn & Real & Syn \\
    \midrule
        Train & 124,864 & 60,850  & 71,108  & 60,850  & 52,981 & -\\
        Dev  &  21,102  &  10,382 & 12,146 & 10,382   & 9,042 &  -\\
        Test &  17,306  &  10,358 & 9,907   & 10,358  & 7,268  & -\\
    % \midrule
        % Total & & & \\
     \bottomrule
    \end{tabular}
    }
    \vspace{-6pt}
    \caption{Number of claim-evidence pairs in \thedataset. Breakdowns of real vs.\ synthetic revisions are presented on the left and right of each cell, respectively.}
    \vspace{-12pt}
    \label{tab:data_stats}
\end{table}

%% file: sections/tasks.tex
\section{\thedataset Tasks} \label{sec:tasks}

\begin{figure*}[!t]
    \centering
    \includegraphics[width=1\textwidth]{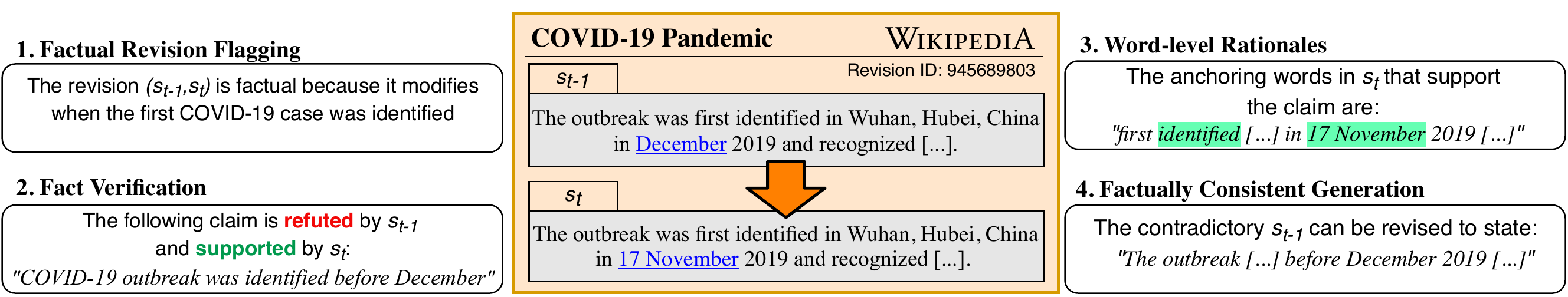}
    \vspace{-16pt}
    \caption{The \thedataset dataset uses Wikipedia revisions to motivate four central fact verification tasks. \\Revision source: \url{https://en.wikipedia.org/wiki/?diff=945689803}.}
    \label{fig:tasks}
        \vspace{-14pt}
\end{figure*}

The unique structure of \thedataset allows us to derive annotations that provide a novel source of supervision for several fact-verification-related tasks. We describe the four main tasks we consider in this work, along with baseline models: (1) factual revision flagging, (2)  fact verification, (3) word-level rationales, and (4) factually consistent generation. Figure~\ref{fig:tasks} illustrates an example from \thedataset. 
We use the following notations:
\begin{itemize}[leftmargin=*, noitemsep]
    \item $\claims$ is the space of  short sentences that express an arbitrary factual statement that can potentially be verified or debunked by external sources.\vspace{3pt}
    \item $\evs$ is the space of sentences that can be found in a trusted online resource (Wikipedia in this study).\vspace{3pt}
    \item $\revtup$ denotes the two versions of a sentence that was revised from $s_{t-1}$ to $s_t \in \evs$.\vspace{3pt}
    \item $\rel(c,s)$ denotes the relation between the claim $c \in \claims$ and observed evidence $s \in \evs$---which can either support $c$ ($\SUP$), refute it ($\REF$), or not contain enough information ($\NEI$).
\end{itemize}

% Next, we provide details and formal definitions for the \thedataset tasks.

\subsection{Factual Revision Flagging} \label{sec:flagging}
Online resources like Wikipedia are continuously changing. In order to remain a reliable and neutral source for recent information, its active community of users must constantly verify and correct the revisions of others. We define \emph{factual revision flagging} as the task of identifying revisions that introduce a \emph{factual} change---e.g., by either modifying a certain fact, adding a new one, or removing an existing one. Such an automated detection process can help the community moderate important articles by serving as a watchdog for factual revisions. Furthermore, tracking factual revisions to certain articles can potentially help keep reliant articles consistent (e.g., citing articles, or non-English versions).

We pose factual revision flagging as a binary classification function $f_{\mathrm{flag}} \colon \evs \times \evs \rightarrow \{0,1\}$, where for a revision $\revtup_i$, we set $y_i=1$ iff there exists a claim in $\claims$ whose label ($\SUP$ or $\REF$) changes as a result of the edit (i.e., $\SUP \rightarrow \{\REF, \NEI\}$ or $\REF \rightarrow \{\SUP, \NEI\}$). Table~\ref{tab:example_revisions} provides example factual and non-factual revisions.
We evaluate the following baseline models:

    \paragraph{Edit Distance.} We measure the edit distance between $s_{t-1}$ and $s_t$, assuming that larger edits are more likely to represent substantive changes. We tune a decision threshold on the validation set.
    
    \paragraph{BOW.} We use an MLP on top of a bag-of-words representation. Each sentence is encoded as $e_*$, the average  \emph{fast}Text~\citep{bojanowski-etal-2017-enriching} word embedding of its edited words (i.e., that were removed or modified in the revision). The MLP input is then taken as $[e_{t-1}; e_{t};|e_{t} - e_{t-1}|; e_{t} \cdot e_{t-1}].$
    
    \paragraph{ALBERT.} We train the ALBERT transformer~\cite{lan-etal-2020-learning} using either only the edited words (diff), or the full sentence pair (full).

\subsection{Fact Verification} \label{sec:fact_ver}
Our basic setting is similar to the inference task of the FEVER dataset.\footnote{To focus on the inference task, as opposed to a full end-to-end system, we assume that we have access to an oracle retriever.} 
We predict the verdict for a claim given an observed evidence, $f_{\mathrm{verdict}} \colon \claims \times \evs \rightarrow \{\SUP, \REF, \NEI\}$.
The FEVER dataset, however, contains independent claim-evidence pairs. In our setting, we have claims paired with revisions such that $\rel(c_i, s_{t-1}) \ne \rel(c_i, s_{t})$, creating contrastive triplets. 
For example, the claim in \figref{fig:tasks} states that the COVID-19 outbreak was identified before December. \thedataset matches it with two different contexts (before and after the presented revision), that can either support or refute that claim.

% We focus on learning the verdict for a claim given a single evidence, $f_{\mathrm{verdict}} \colon \claims \times \evs \rightarrow \{\SUP, \REF, \NEI\}$. We note, however, that an extension to multiple evidence sentences is possible by either concatenating the inputs or aggregating the predictions.
% Contextual fact verification extends the standard fact verification task by requiring the verdict of the classifier to depend on the presented evidence. This property is important for (1) utilizing dynamically changing evidence sources that provide the most recent information, and (2) explaining the verdict by including the responsible evidence.

% Our basic setting is similar to the inference task of the FEVER dataset.\footnote{To focus on the inference task, as opposed to a full end-to-end system, we assume that we have access to an oracle retriever.} FEVER, however, is based on a single timestamp of Wikipedia. In our setting, we have claims paired with revisions such that $\rel(c_i, s_{t-1}) \ne \rel(c_i, s_{t})$, ensuring that the contextual element of the task must be upheld. We focus on learning the verdict for a claim given a single evidence, $f_{\mathrm{verdict}} \colon \claims \times \evs \rightarrow \{\SUP, \REF, \NEI\}$. We note, however, that an extension to multiple evidence sentences is possible by either concatenating the inputs or aggregating the predictions.

\label{sec:fact-model}
Our baseline model is an ALBERT sentence-pair classifier that predicts $\rel(c,s)$. Compared to BERT~\citep{devlin-etal-2019-bert}, it uses fewer parameters by shrinking the embedding size and sharing layers, which we find to improve robustness.

\subsection{Word-level Rationales} \label{sec:rationale}
Word-level rationales  provide useful explanations for predictions of neural models~\citep{lei-etal-2016-rationalizing}. Such explanations can be particularly useful for semi-automated fact verification, since they allow users to quickly interpret and trust the model's verdict.\footnote{ \citet{covid-infodemic} showed that explanations can increase the agreement between users and expert fact-checkers.}
In \figref{fig:tasks}, for example, the date of the first identified case can explain the verdict for the claim. 

As first proposed by \citet{lei-etal-2016-rationalizing}, the standard definition of extractive rationales asks for selecting the minimal set of input tokens that is sufficient for preserving the model's prediction. Here we use a slightly modified definition following \citet{shah2020automatic}, where we identify the minimal set of evidence tokens where removing them will change the input's label to $\NEI$.

We pose this task as conditional masking, where we learn a function $f_{\mathrm{rationale}} \colon \claims \times \evs \rightarrow \{0, 1\}^n$, where $n$ is the length of an evidence $s \in \evs$.  Given an evidence $s=(x_1, \dots, x_n)$ and a claim $c$, where $\rel(c,s) \in \{\SUP, \REF\}$, we want to find a mask $m$ such that $\rel(c, s \odot m) = \NEI$, where
\begin{equation*}
  s \odot m =
    \begin{cases}
      x_i & \text{if } m[i] = 0;\\
      \texttt{<mask>} & \text{if } m[i] = 1.
    \end{cases}  
    \vspace{-2pt}
\end{equation*}

Moreover, we want $m$ to be as sparse as possible. Intuitively, $s \odot m$ could be viewed as an incomplete revision in which the masked words that have not yet been filled in will determine the relation with the claim. We say that $m$ reveals the most responsible words in $s$ for resolving $c$. Following \citet{shah2020automatic}, we formulate an unsupervised objective as
\begin{equation}
\label{eq:rationale}
   \min \sum_{i=1}^n m_i ~\text{  s.t.  }~ \rel(c,s \odot m) = \NEI.  
   \vspace{-4pt}
\end{equation}

We evaluate the quality of $m$ by comparing it in terms of F1 to both (1) $m_{\mathrm{edit}}$, the non-stopwords removed or replaced in the true revision (i.e., \emph{edit prediction}), and (2) $m_{\mathrm{manual}}$, a manually annotated ``human'' reference, (i.e., \emph{rationale prediction}).
We implement the following two baselines:

\paragraph{Unsupervised.} As in \citet{shah2020automatic}, we optimize a Lagrangian relaxation of Eq.~\ref{eq:rationale}, where
\[
\mathcal{L}_{\mathrm{us}} := -\log p(\rel(c, s \odot m) = \NEI) + \frac{\lambda}{n} \sum_{i=1}^n m_i.
%\vspace{-2pt}
\]
We keep the $\rel$ classifier (from \S\ref{sec:fact-model}) fixed, and train a separate ALBERT model to predict the mask $m$ using a Gumbel softmax~\cite{gumbel}.

\paragraph{Distantly Supervised.} By leveraging opposing claims present in \thedataset, we are able to identify $m_{\mathrm{edit}} = \texttt{diff}(s_{t-1}, s_t)$---i.e., the non-stopwords that are deleted or replaced in $s_{t-1}$ when compared to $s_t$. We then use $m_{\mathrm{edit}}$ as distant supervision for $m$, where
$\mathcal{L}_{\mathrm{ds}} = -\frac{\gamma}{n}\sum_{i=1}^n \log p(m_i = m_{\mathrm{edit}_i})$. We combine both the $\mathcal{L}_{\mathrm{us}}$ and $\mathcal{L}_{\mathrm{ds}}$ losses.

\subsection{Factually Consistent Generation} \label{sec:generation}
As facts change, the sources reporting them must change as well to reflect the most recent information. In \thedataset, this is reflected via the active revisions to Wikipedia. We simulate \emph{automating} this process by considering two generation tasks:

\paragraph{Automatic Revisions.} Given an outdated context $s_{t-1}$ and an updated claim $c$, we learn $f_{\mathrm{revise}} \colon \evs \times \claims \rightarrow \evs$ to produce a new context $s_{t}$ that minimally modifies $s_{t-1}$ to agree with $c$. 
For example, one can change $s_{t-1}$ in \figref{fig:tasks} to state ``\emph{before December}'' in order to agree with the claim.

\paragraph{Claim Extraction.} Given a revision $\revtup$,  we learn $f_{\mathrm{extract}} \colon \evs \times \evs \rightarrow \claims$ to produce a short claim $c$ that expresses the factual change.

 In both tasks, the output should satisfy $\rel(c, s_{t}) = \SUP$, while $\rel(c, s_{t-1}) = \REF$. We use $f_{\mathrm{verdict}}$ (\S\ref{sec:fact_ver}) to evaluate this requirement.
We experiment with both BART-base~\citep{lewis-etal-2020-bart} and T5-base~\citep{Raffel2020ExploringTL} sequence-to-sequence transformer-based generators. For the revision task, we concatenate $s_{t-1}$ and $c$ with a separator and train the model to predict $s_t$. For the claim extraction task, we combine the input pair $\revtup$ into a single sentence that visualizes the revision (e.g., \textit{``sales of \{4.7 $\rightarrow$ 5.4\} million''}).

% where, given an outdated context $s_{t-1}$ and an updated claim $c$, we train a model $f_{\mathrm{generate}} \colon \evs \times \claims \rightarrow \evs$ to produce a new context $s_{t}$ that is now consistent with $c$. More specifically, if $\rel(c, s_{t-1}) = \REF$, then we want to produce $s_t$ such that $\rel(c, s_{t}) = \SUP$.

%% file: sections/Experiments.tex
\section{Experiments} \label{sec:experiments}

We present and analyze results for the models described in \secref{sec:tasks}. Our analysis attempts to evaluate several questions:
(1) How well can the current state-of-the-art models perform on the \thedataset tasks? (2) Does \thedataset increases the robustness of models against adversarial examples? (3) Can \thedataset improve interpretability by providing supervision for anchoring words?

% \paragraph{Experimental Setting.}
% We implement all our models with the HuggingFace Transformers library~\citep{Wolf2019HuggingFacesTS}. When comparing across training datasets with different size, we train the model for the same amount of update steps, upsampling the smaller datasets. We pick the checkpoint with the highest accuracy on the development set of the training task and report performance on the test sets. For more details see \secref{appendix}.

% We implement all our models with the HuggingFace Transformers library~\citep{Wolf2019HuggingFacesTS}. We fine tune the classifier for 12,000 and 50,000 training steps with with a batch size of 64 and 32 for the flagging and fact verification tasks, respectively. We store checkpoints of the model in fixed intervals during the training and select the one with the highest accuracy on the development set of the training task. Thereafter, we evaluate and report the performance of the selected model on the test set of the relevant tasks.

\subsection{Related Datasets}
In addition to \thedataset, we train and evaluate on several related datasets, which we briefly describe:
% In addition to \thedataset, we train and evaluate on several related datasets, as described below. Models are trained for the same amount of updates, adding epochs for small datasets.

\paragraph{FEVER}~\citep{thorne-etal-2018-fever}: A popular fact verification dataset based on Wikipedia. We use the provided $\SUP$ and $\REF$ claim-evidence pairs. For $\NEI$ claims, we randomly sample neutral evidence from the article with the highest BM25 score.
% As the labels from the shared task test set are hidden, we use the dev and test sets from their paper.

\paragraph{MNLI}~\citep{williams-etal-2018-broad}: A large and diverse dataset for natural language inference. The three-way sentence-pair entailment prediction is similar to fact verification.
    %Each Annotator was asked to create three hypotheses with different relations for each premise, making it closer to our fully symmetric design---but doesn't avoid hypothesis-only bias~\citep{poliak-etal-2018-hypothesis}. 
    We use the hypothesis as the claim and the premise as the evidence and evaluate on the ``mismatched'' evaluation set.
    % \footnote{We evaluate on the MNLI ``mismatched'' setting.}

\paragraph{Symmetric}~\citep{schuster-etal-2019-towards}: A set of challenging symmetric, synthetic extensions to FEVER's evaluation set that avoid claim-only bias.

\paragraph{Adversarial}~\citep{thorne-etal-2019-fever2}: Adversarial examples created by participants of the FEVER 2.0 shared task. Teams were asked to create claims that break FEVER-trained models. We take all $\SUP$ and $\REF$ claims and their gold evidence sentences.

\paragraph{Triggers}~\citep{atanasova-etal-2020-generating}: A set of 186 FEVER claims paraphrased adversarially to contain universal adversarial triggers~\citep{wallace-etal-2019-universal}. Its small size leads to high variance results. 

\paragraph{ANLI}~\citep{nie-etal-2020-adversarial}: An adversarial dataset for MNLI- and FEVER-based models. The creation was performed in three iterative rounds in which a model was trained, and then  crowdworkers devised adversarial inputs, and the process repeated.

\paragraph{PAWS}~\citep{zhang-etal-2019-paws}: A dataset of altered Wikipedia sentences using word swapping and back-translation. Human annotators labeled whether the modified sentence is a paraphrase or not. 
We evaluate whether a PAWS-trained classifier can be used for our factual revision flagging task.

\subsection{Factual Revision Flagging} \label{sec:expr_flagging}

\input{tables/res_flag}

\tabref{tab:res_flag} shows the results of our baseline models on the factual revision flagging task. First, we notice that a model trained on the PAWS dataset (reaching 93.42 F1 score on PAWS test) does \emph{not} transfer well to the flagging task, and performs on par with a simple edit distance heuristic. We hypothesize that this is a result of the entity scrambling technique used to synthetically revise sentences in PAWS, which is different from the edits introduced by real, factual Wikipedia revisions in practice. 

Second, we see that the performance of neural models trained on the $\thedataset$ flagging task increases with richer inputs and more advanced models---demonstrating the complexity of the task. The ALBERT (diff) model that uses only the modified word sequences from each sentence (i.e., contextual within a subspan) improves the AUC by 10 points over a BOW model that gets a similar input. The ALBERT (full) model that receives the full sentences as input (i.e., has access to even more context), further improves the AUC by 2 points. Nevertheless, the best model still only reaches 83 macro-F1, indicating the difficulty of this task.

\subsection{Fact Verification} \label{sec:exp_fact_ver}

\input{tables/res_fact_ver}

\begin{figure}[t]
    \centering
    \includegraphics[width=0.5\textwidth, trim=0 0 0 31, clip]{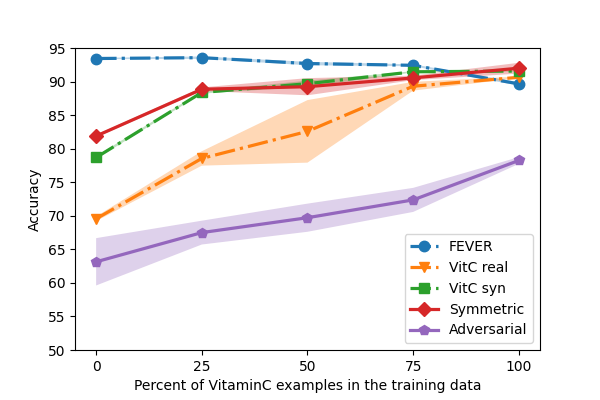}
    \caption{Test accuracy of models trained on a dataset of 100K combined $\SUP$ and $\REF$ examples from $\thedataset$ and FEVER. The higher the ratio of $\thedataset$ in the training data, the better the performance on adversarial evaluation sets (solid lines). The shaded areas represent standard error across three runs.}
    \label{fig:break_fact_ver}
    \vspace{-6pt}
\end{figure}

\tabref{tab:res_fact_ver} summarizes the results for classifiers trained on fact verification and NLI datasets. Verifying claims against real revisions proves to be the hardest. The best model achieves 89\% accuracy, lower than that on either \thedataset's synthetic cases or the original FEVER examples. %The best model achieves 89\% accuracy, which is similar to the in-domain performance on MNLI (that is known to be challenging). Augmenting FEVER data with MNLI, however, has limited effect on adversarial examples. 
%
%This demonstrates the effectiveness of our dataset. 
Including \thedataset examples in the training data drastically increases models' sensitivity to contrastive examples (rightmost column)---while preserving the in-domain accuracy (only $-0.42\%$ for FEVER and $+0.12\%$ for MNLI with ALBERT-xlarge).
Another evidence for the generalization properties conferred by  \thedataset is its zero-shot performance to both other datsets. An ALBERT-xlarge model trained only on \thedataset reaches $76\%$ and $79\%$ accuracy on FEVER and MNLI, respectively. In contrast, the transfer accuracy for MNLI$\rightarrow$FEVER is $70\%$ and for FEVER$\rightarrow$MNLI is only $38\%$.

Most importantly, models trained with \thedataset perform better on challenging adversarial datasets. On the otherhand, simply augmenting FEVER data with MNLI data has a limited effect on adversarial examples.\footnote{We've also tried augmenting FEVER with ANLI for an ALBERT-xlarge model and find it to achieve only $73\%$, $91\%$, and $34\%$ on Adver., Sym., and Triggers, respectively.} We conjecture that the contrastive nature of \thedataset helps models better learn the relations between the claims and evidences---and to avoid relying on certain artifacts that do not generalize well.
 
To further probe the value of \thedataset  examples compared to FEVER ones ($\SUP$ and $\REF$ only), we compose training sets of 100K examples using different ratios of the two datasets. As shown in \figref{fig:break_fact_ver}, including more \thedataset pairs continuously improves the performance on the challenging adversarial and symmetric evaluation sets.

As an additional qualitative experiment, given the recent successes of huge language models such as GPT-3~\citep{brown2020language}, we explore whether such models develop sufficient context sensitivity on their own. Appendix~\ref{app:gpt} shows the results of classifying several claims using a few-shot GPT-3 model. We find that GPT-3 still largely under-performs our \thedataset-trained models in terms of sensitivity---demonstrating the importance of using \thedataset's unique structure during training.

\subsection{Word-level Rationales} \label{sec:exp_rationale}
\input{tables/res_rationale}
\tabref{tab:res_rationale} shows the results of our baseline models for identifying word-level rationales (i.e., anchoring words in the evidence). While our unsupervised model is able to uncover some patterns, directly leveraging the structure of \thedataset to obtain distant supervision for likely anchoring words (i.e., token labels) improves both the edit prediction and the word-level rationale prediction performance.\footnote{We evaluate rationales using a manually annotated test set of 300 examples (150 each from VitC real and VitC synthetic).}
Example predictions are provided in Appendix~\ref{app:examples}.

\subsection{Factually Consistent Generation} \label{sec:exp_generation}

\input{tables/res_generation}

\begin{table*}[h!]
\centering
\small
    \resizebox{2.05\columnwidth}{!}{%
%   \begin{tabular}{p{1.8cm}|p{13.5cm}}
\begin{tabular}{p{1.8cm}|p{12.5cm}|p{0.5cm}}
%\multicolumn{3}{r}{$f_{\mathrm{verdict}}(c,s_t)$}  \\
\toprule

$\revtup$    & \multicolumn{2}{l}{2020 coronavirus pandemic in Germany: there have been\{*| 349 -> 444 |\}confirmed cases and 16 recoveries.} \\
\cmidrule{2-3} 
BART (VitC) &  \multicolumn{1}{l}{More than 400 people have tested positive for COVID-19 in Germany. \qquad \qquad \qquad $\vert$ \  $f_{\mathrm{verdict}}(c,s_t)=$} & $\SUP$ \\
\cmidrule{2-2} 
GPT-3 $\mathcal{T}$=$0$ & As of 14 March, there have been more than 350 confirmed cases of the virus in Germany & $\SUP$\\
\cmidrule{2-2} 
GPT-3 $\mathcal{T}$=$0.7$ & As of March 12, there have been more than 400 confirmed cases and 20 reported deaths & $\NEI$\\
\cmidrule{2-2} 
Reference & \textcolor{darkgreen}{There have been more than 400 confirmed coronavirus cases in Germany .}& $\SUP$\\
%\bottomrule
\midrule
$\revtup$    & \multicolumn{2}{l}{\multirowcell{2}[0ex][l]{Diego Corrales: Corrales was born to a\{*| Puerto Rican -> African American |\}father and a\\\{*| Dominican -> Mexican |\}mother .}}\\
& \multicolumn{2}{c}{}\\
\cmidrule{2-3} 
BART (VitC) &  Diego Corrales'father is African American and his mother is Mexican. & $\SUP$\\
\cmidrule{2-2} 
GPT-3 $\mathcal{T}$=$0$ & Corrales was born to a Puerto Rican father and a Mexican mother & $\REF$\\
\cmidrule{2-2} 
GPT-3 $\mathcal{T}$=$0.7$ & Corrales was born to a father from Puerto Rico and a mother from the Dominican Republic & $\REF$\\
\cmidrule{2-2} 
Reference & \textcolor{darkgreen}{Diego Corrales´ father was African American and his mother Mexican .}& $\SUP$\\

\bottomrule
\end{tabular}
}%
\caption{Example outputs for expressing claims that reflect the factual changes in a single Wikipedia revision. The BART-base model is trained on \thedataset data, while GPT-3 is applied in a 2-shot setting with a temperature of 0 or 0.7 (see Appendix~\ref{app:gpt}). The revision $\revtup$ is given to the model as a single sentence, where the edits are between curly brackets. The human-written claim is provided for reference. The rightmost column contains the prediction of our ALBERT-xlarge $f_{\mathrm{verdict}}(c,s_t)$ model (trained on \thedataset) when using the generated claim.} \label{tab:gen_claim_out}
    \vspace{-10pt}
\end{table*}

%  We use several evaluation metrics to assess the quality of the outputs.
\tabref{tab:res_gen} presents the results on factually consistent generation. We find BART to perform better in both of our generation tasks (though we only tried the default setting). The BLEU score~\citep{papineni-etal-2002-bleu} is lower in the claim extraction task since there is freedom in how to phrase the claims, which can result in greater differences between the outputs and the references. The BERT-based BLEURT score~\citep{sellam-etal-2020-bleurt} shows a similar trend. Still, the claim extraction model succeeds in updating the facts that reflect the true revision $86\%$ of the time, as measured by the fact verification model's verdict ($f_{\mathrm{verdict}}$).

The revision generator aims to modify sentences so that they agree with a given claim. According to our fact verification model's verdict, it succeeds in doing so 76\% of the time. Furthermore, revisions should resemble real ones, and preserve the remaining content that is unrelated to the claim. The SARI KEEP F1~\citep{xu-etal-2016-optimizing} of 75 shows that the model and the reference mostly agree on parts of the sentence that should be kept unchanged.
% The claim extraction model succeeds in distilling the facts that reflect the revision for 86\% of the cases.

We find that the token-based measures and our $f_{\mathrm{verdict}}$ metric agree well with human (manual) evaluation scores. We randomly sampled 100 generated and human-written sentences per task, and asked workers on Amazon MTurk to rate their grammaticality and whether the evidence $s_t$ supports the claim. The scores of the generated sentences were on par with the human-written ones, indicating the high-quality of our outputs.

% \textcolor{red}{TODO: gpt3}

\tabref{tab:gen_claim_out} presents two example generations for the claim extraction task (we provide additional qualitative examples in Appendix~\ref{app:examples}). Our model is able to efficiently extract a self-contained claim that expresses the correct fact after the edit.
As in \S\ref{sec:exp_fact_ver}, we also explore how GPT-3 handles this task (we provide two demonstrations in the prompt). Compared to the BART model trained on \thedataset, GPT-3 appears to make more factually inconsistent or unsupported generations (see Appendix~\ref{app:gpt} for more details). Encouragingly, our  $f_{\mathrm{verdict}}$ classifier is still able to pick up on this---as demonstrated by the predictions in the rightmost column of Table~\ref{tab:gen_claim_out}. 
For example, classifying the report about 20 deaths as $\NEI$ since it is not part of the source.
Once again, this serves to qualitatively demonstrate the effectiveness of leveraging \thedataset.
%We also experiment with a two-shot GPT-3 model (see Appendix~\ref{app:gpt} for more details). While only two guiding examples allow it to generate a focused claim, the output can include made-up facts with no supporting evidence (especially when $\mathcal{T}>0$). For example, it reports about 20 deaths although no such thing was mentioned in the input.

%% file: tables/res_flag.tex
\begin{table}[t]
\centering
    \resizebox{\columnwidth}{!}{%
    \small
\begin{tabular}{ll|cccc}
\toprule
Model    &   Train data  & AUC  & Prec. & Rec. & F1    \\
\midrule
Edit dist. & - & 71.34 & 64.90 & 63.18 & 63.56    \\
ALBERT & PAWS-full & 72.20 & 65.27 & 60.61 & 60.48 \\
\midrule
BOW          & VitC-diff & 79.87 & 70.85 & 67.84 & 68.55 \\
ALBERT & VitC-diff & 89.87 & 80.69 & 82.06 & 81.18 \\
ALBERT  & VitC-full & \textbf{91.97} & 82.63 & 84.49 & \textbf{83.18} \\
\bottomrule
\end{tabular}
    }
    \caption{Factual revision flagging scores for models aware of the full sentence-pair (full) and aware only of the modified words (diff). We use ALBERT-base.} \label{tab:res_flag}
   \vspace{-10pt}
\end{table}

%% file: tables/res_fact_ver.tex
\begin{table*}[t]
\centering
\small
\resizebox{2\columnwidth}{!}{%
\begin{tabular}{l|l|cccc|cccc|c}
\toprule
Model & Train dataset & VitC real & VitC syn & FEVER & MNLI &  Adver. & Sym.  & Triggers  & ANLI & Contrast \\
\midrule
\multirow{6}{1cm}{ALBERT-base} 

& FEVER                & 54.78   & 79.18  & 95.07 &   58.45   & 62.01  & 81.18 &  3.33    & 32.94  & 55.53 \\
& MNLI &             44.93 &	69.67 &	65.70 &	85.22 &	49.61 &	72.89 &	68.82  &	30.63  & 56.67 \\
& FEVER + MNLI &     55.93 &	82.26 &	95.62 &	85.58 &	63.97 &	85.67 &	38.71  &	30.59 & 61.90 \\
\cmidrule{2-11}
& VitC                 & 86.16   & 89.78  & 74.56 &   69.18   & 71.41  & 90.17 &  \textbf{86.67}   & 37.25     & 84.11   \\
& VitC + MNLI &      86.68 &	91.18 &	77.70 &	85.95 &	69.58 &	91.15 &	70.97  &	34.31 & 85.61 \\

& VitC + FEVER         & 86.26   & 91.05  & 94.24  &   68.90  & 68.80  & 91.57 & 72.04   & 38.50   & 85.89 \\
\midrule

% \multirow{6}{*}{BERT-base}

% & FEVER                &  60.55 & 71.35 &	87.16 &	61.90 &	52.09 &	73.60 &	69.89  &	\textbf{34.53}         \\
% & MNLI &           46.31 &	69.01 &	70.06 &	83.80 &	50.13 &	73.88 &	65.05  &	26.88 \\
% & FEVER + MNLI &   56.24 &	81.80 &	\textbf{95.59} &	\textbf{85.06} &	63.05 &	85.11 &	37.63 &	29.63 \\
% \cmidrule{2-10}
% & VitD                 &  \textbf{85.80} & 90.63 &	74.21 &	66.66 &	\textbf{76.24} &	90.17 &	63.98 &	33.19         \\
% & VitD + MNLI &    84.47 &	\textbf{91.00} &	74.88 &	83.70 &	63.05 &	84.55 &	66.13 &	31.00 \\
% & VitD + FEVER         &  84.72 & 89.16 &	87.55 &	69.28 &	64.75 &	\textbf{90.73} &	\textbf{72.58}  &	34.06           \\

% \midrule

\multirow{6}{1.5cm}{ALBERT-xlarge}
& FEVER         & 58.56 & 84.22 & 96.47 & 38.33 &	72.58 &	87.08 &	32.80  &	36.03 & 64.59 \\
& MNLI          & 49.26 & 74.58 & 70.47 & 88.91 &	53.52 &	78.65 &	72.58  &	36.38 & 63.41 \\ 
& FEVER + MNLI  & 61.21 & 86.81 & \textbf{96.81} & \textbf{89.04} &	69.97 &	89.75 &	43.55  &	37.66 &  70.44\\
\cmidrule{2-11} 
& VitC          & 88.64 & 93.84 & 75.99 & 78.89 &	\textbf{82.51} &	\textbf{94.80} &	67.20  &	\textbf{42.66} & 89.57 \\
& VitC + MNLI   & 88.69 & 93.92 & 76.80 & 89.03 &	80.81 &	\textbf{94.80} &	67.20  &	40.09 & 89.80 \\ 
& VitC + FEVER  & \textbf{88.87} & \textbf{94.01} & 96.05 & 62.30 &	80.94 &	\textbf{94.80} &	65.59  &	40.31 & \textbf{90.94} \\ 
\bottomrule

\end{tabular}
}%
\caption{Test accuracy of fact verification and NLI models. \thedataset-trained models are more robust to adversarial examples and more sensitive to contrastive contexts. The rightmost column shows the percent of FEVER claims in which the prediction flipped when presented with contrastive contexts.} \label{tab:res_fact_ver}
\vspace{-14pt}
\end{table*}

%% file: tables/res_rationale.tex
% \begin{table}[t]
% \centering
% \small
% \begin{tabular}{l|ccc}
% \toprule
% Model                & F1    & Recall & Precision \\ \midrule
% Unsupervised         & 37.33 & 46.64  & 31.11     \\
% Distantly supervised & \textbf{47.36} & \textbf{66.03}  & \textbf{36.92}     \\ \bottomrule
% \end{tabular}

% \end{table}

% \usepackage{booktabs}
\begin{table}[t]
\small
    \resizebox{1\columnwidth}{!}{%
\begin{tabular}{c|ccc|ccc}
\toprule
Token & \multicolumn{3}{c}{\ul{Edit prediction}} & \multicolumn{3}{c}{\ul{Word-level rationales}} \\
labels        & Prec.     & Rec.    & F1   & Prec.     & Rec.    & F1    \\ \midrule
- & 31.93     & 30.56  & 31.23          & 31.11 & 46.64  & 37.33 \\
$\checkmark$ & 55.93     & 62.68  & \textbf{59.11}      & 36.92 & 66.03  & \textbf{47.36} \\ \bottomrule
\end{tabular}
}
\caption{The distant token-level supervision of \thedataset improves the edit prediction, and as result identifies the anchoring words (rationales) more accurately.}
% improves the sensitivity to identifying polarizing word level rationales.}
\vspace{-12pt}
\label{tab:res_rationale}
\end{table}

% Model        & Rationale  & F1    & Recall & Precision \\ \midrule
% Unsupervised & Revision   & 31.23 & 30.56  & 31.93     \\
% Distant sup. & Revision   & 59.11 & 62.68  & 55.93     \\ \midrule
% Unsupervised & Manual     & 37.33 & 46.64  & 31.11     \\
% Distant sup. & Manual     & 47.36 & 66.03  & 36.92     \\ \bottomrule

%% file: tables/res_generation.tex
\begin{table*}[h!]
\centering
\small
    \resizebox{2\columnwidth}{!}{%
\begin{tabular}{cc|cc|cccc|c|c|cc}
\toprule
% Target & Model   	   & BLEU & KEEP  & SARI & BLEURT & $f_{\mathrm{verdict}}$ \\
% \midrule
% \multirow{2}{*}{Revision} & T5	    	      & 47.46 & 72.61  & 42.99 & 0.38 & 64.52 \\
%  & BART                                       & 54.86 & 75.36  & \textbf{47.21} & 0.67 & \textbf{76.26} \\
% \midrule
% \multirow{2}{*}{Claim} & T5 & 13.95 & 44.36 & 50.83 & -0.12 & 75.39  \\
% & BART &                    16.14 & 52.91 & \textbf{55.97} & 0.16 & \textbf{85.83} \\

Target & Model & \multicolumn{2}{c|}{} & \multicolumn{4}{c|}{\ul{SARI scores}} & & & \multicolumn{2}{c}{\ul{Manual evaluation}}\\
  &	  & $\mathrm{ROUGE}_2$ & BLEU & KEEP & ADD & DEL & AVG & BLEURT & $f_{\mathrm{verdict}}$ & Grammar & SUP \\
\midrule
\multirow{2}{*}{Revision} & T5   & 77.63 & 47.46 & 72.61	& 13.32 & 43.04 & 42.99 & 0.38 & 64.52 & 81.00 & 71.80\\
 & BART   	 & 85.23 & 54.86 & 75.36	& 18.31 & 47.95 & 47.21 & 0.67 & 76.26 & 84.80 &	83.20\\
\midrule
\multirow{2}{*}{Claim} & T5  & 35.19 & 13.95 & 44.36 & 20.59 & 87.54 & 50.83 & -0.12 & 75.39 & 71.33 &	72.22  \\
& BART  & 40.38 & 16.14 & 52.91 & 23.62 & 91.37 & 55.97 & 0.16 & 85.83 & 75.78 &	74.22 \\

\bottomrule

\end{tabular}
    }
    \caption{Factually consistent generation results. Higher is better for all scores and the max value is $100$ (except for BLEURT). $f_{\mathrm{verdict}}$ is the score of our \thedataset-trained ALBERT-base model on the outputs. For manual evaluation, outputs were rated by their grammaticality and by how much the evidence supports the claim (SUP). For reference, human-written pairs received $75.75$ and $76.0$ average scores for Grammar and SUP, respectively.} \label{tab:res_gen}
    %\vspace{-6pt}
\end{table*}

%% file: sections/discussion.tex
\section{Conclusion} \label{sec:discussion}

We presented \thedataset, a large-scale dataset for training and evaluating fact verification models using contrastive contexts. Our novel method of leveraging factual revisions to Wikipedia enabled us to create challenging examples in which a claim is paired with contexts that are lexically similar, yet factually opposing. Our results illustrated that training on \thedataset improves classifier sensitivity to subtle changes in evidence, and increases their robustness to adversarial examples.

Furthermore, we formulated several new, important tasks for fact verification that \thedataset allows us to test. We showed how the dataset's unique ``before and after'' structure lends itself to training classifiers to flag factual revisions. In addition, for factual revisions, the edits reveal which words in the evidence are the most critical---which helps supervise word-level rationale models for better interpretability. Finally, we demonstrated that \thedataset can help with factually consistent text generation. We hope that this work and the range of tasks it presents will motivate and support the fact verification field in developing reliable models that can adapt to dynamically changing evidence.

% We present a large-scale dataset for training and evaluating fact verification models. We base the claim-evidence pairs on Wikipedia revisions, thereby creating contrastive triplets of a claim and two lexically similar but factually different contexts that reflect opposite relations with the claim. We show that the training examples from our dataset improve the sensitivity of classifiers to contrastive evidence and increases the robustness against adversarial examples.

% In addition, our dataset improves the explainability of fact verification models by ensuring that the verdict is conditioned on the retrieved sentence, and helps in identifying the word-level rationales in the evidence.
% Finally, we use the data to advance the automation of other related tasks by introducing a factual revision flagger and factually consistent text generators. The flagger can help content moderators identify check-worthy revisions or keep track of edits in related articles, and the generator can distill the factual information from the revision or provide edit suggestions to articles from given claims.

%% file: sections/acknolegments.tex
\section*{Acknowledgements}
% We thank the TransPerfect team for their cooperativeness, and Darsh J Shah and Enrico Santus for helpful discussions.
We thank the TransPerfect team, Darsh J Shah and Enrico Santus for helpful discussions, as well as the members of the MIT NLP group and Andreas Vlachos for valuable feedback.  
% We thank OpenAI for providing beta access to GPT-3. 
This work is supported in part by the Facebook Online Safety Benchmark Award. TS is supported by in part by DSO grant DSOCL18002. AF is supported in part by a NSF Graduate Research Fellowship.

%% file: appendix/data_stats.tex
% \begin{center}
% {\bf \Large{Appendix}}
% \end{center}
\counterwithin{figure}{section}
\counterwithin{table}{section}

\section{\thedataset: Complementary details}
\label{app:stats}

We provide additional details about the \thedataset dataset. 
% Examples of cases from the dataset are available in \tabref{tab:example_revisions} along with model's predictions in \appref{app:examples}.

% \begin{table}[h]
% \centering
% \small
%   \begin{tabular}{p{1cm}|p{6cm}|p{7cm}}
% \toprule
% Relation &  claim & Evidence      \\
% \midrule

% $\SUP$    &   &  \\
% \cmidrule{2-3} 
%  & &\\

% \midrule

% \bottomrule
% \end{tabular}
% \caption{Examples of claim-evidence pairs from the \thedataset real} \label{tab:vitc_real_cases}
% \end{table}

\subsection{Claim Statistics}

\begin{figure}[h]
    \centering
    \includegraphics[width=.4\textwidth]{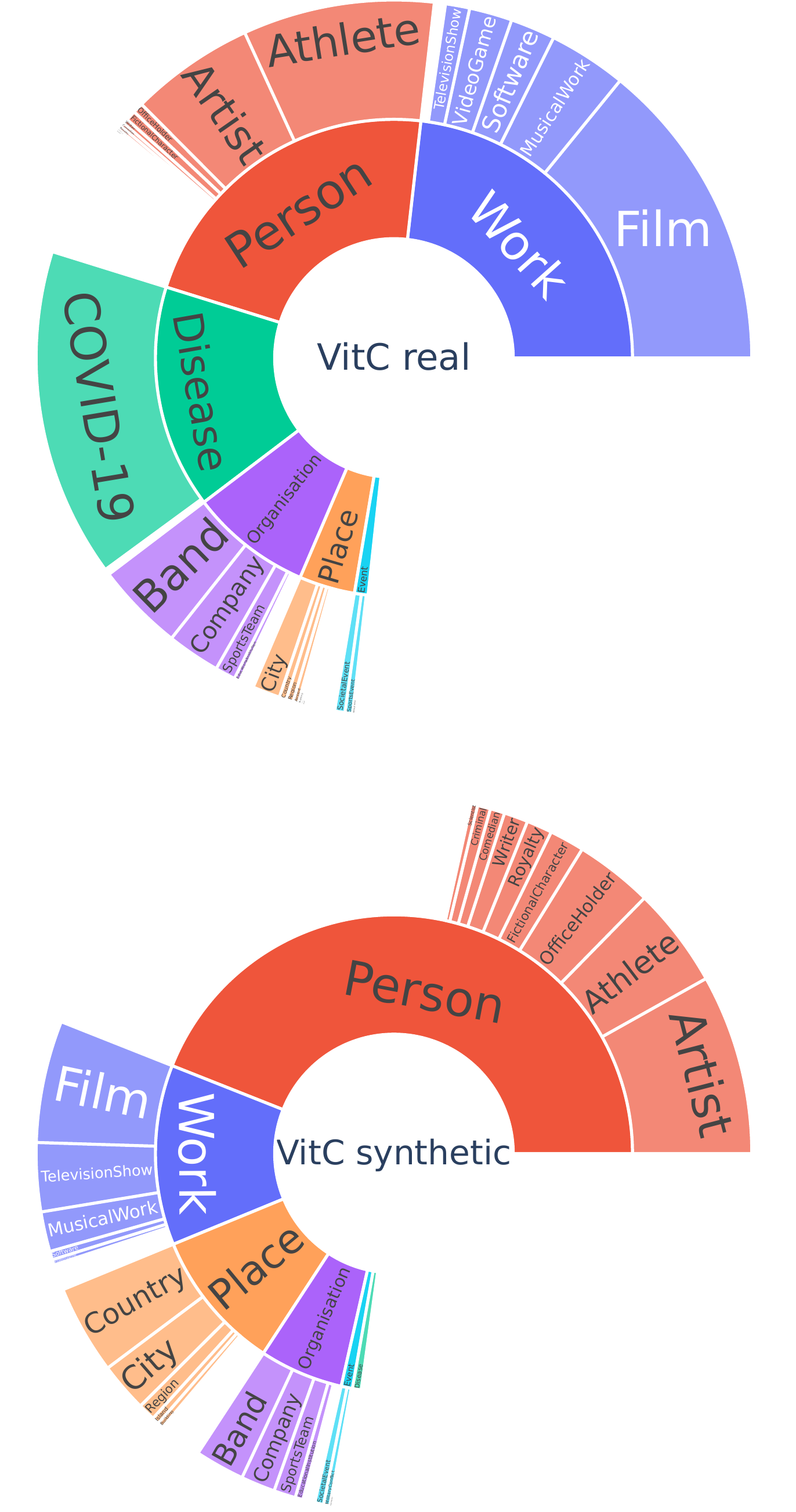}
    \caption{Distribution of claims in the $\thedataset$ dataset by the topic of the originated Wikipedia article.}
    
    \label{fig:topics}
\end{figure}

\paragraph{Topic Distribution.}
\figref{fig:topics} shows the distribution of claims in the \thedataset dataset by the topic of the Wikipedia article they are based on. The information was collected from DBpedia,\footnote{\url{http://dbpedia.org/ontology/}} retrieving the parent class of the pages. Labels for about 25\% of the articles were missing, and left blank in the diagram.

The ``synthetic'' part of \thedataset, which is based on the claims of the FEVER dataset, contains many claims about specific human entities. About 15\% of the claims in \thedataset real are about COVID-19.

\begin{table*}[h]
\centering
\small
% \resizebox{1\columnwidth}{!}{%
\begin{tabular}{l|r|r| l}
\toprule
Claim category &  real & synthetic & Example  \\

\midrule

Quantitative & 48\% & 9\% & The COVID-19 pathogen may last less than 10 days on some surfaces.\\
Calendrical & 9\% & 15\% & Italy surpassed the 10,000 coronavirus-related deaths on a Saturday.\\
% A statement regarding a location & 2\% & 4\% \\ 
Entity & 23\% & 58\% & Mary of Teck was queen-consort.\\
Event & 14\% & 14\% & In the last EFL Cup, Manchester defeated Chelsea.\\ 
Other & 6\% & 4\% & Most genes need further research to better understand the function of their RNA products. \\

\bottomrule

\end{tabular}
% }%
\caption{Estimated distribution of claims in the $\thedataset$, based on manual annotations of 100 randomly sampled claims from the development split of the real and synthetic subsets. An example claim from each category is provided for reference.}  \label{tab:claim_cat}

\end{table*}

\paragraph{Category Distribution.} \label{sec:claim_cat}
We sample 100 examples from the ``real'' and ``synthetic''  subsets of \thedataset and manually categorize their claims. Due to the creation methodology of \thedataset real, its claims mostly describe frequently updating facts, or facts that tend to be corrected. We find about half of these claims to describe changes in numerical values (e.g., number of COVID-19 cases, earnings or ratings of movies, number of awards etc.). In contrast, \thedataset synthetic mostly covers general facts about specific entities, (e.g., place of birth, date of birth, occupation, etc.). This is a result of the synthetic claims being based on the FEVER dataset, where annotators were asked to come up with claims on popular Wikipedia pages. Combined, the \thedataset dataset holds a diverse set of claims about various topics.

\subsection{Inter-annotator Agreement}
We ask three additional annotators to independently annotate a random set of two thousand claim-evidence pairs, evenly distributed between the development and test splits of the real and synthetic sets. The Fleiss $\kappa$ score~\citep{fleiss1971} between the four annotations is 0.7065, which means substantial agreement. Similar agreement scores of 0.6841 and 0.7 were reported for fact verification~\citep{thorne-etal-2018-fever} and NLI datasets~\citep{bowman-etal-2015-large}, respectively.

\begin{figure*}[h]
\small
\centering
\begin{subfigure}{0.32\textwidth}

\includegraphics[width=1.05\linewidth]{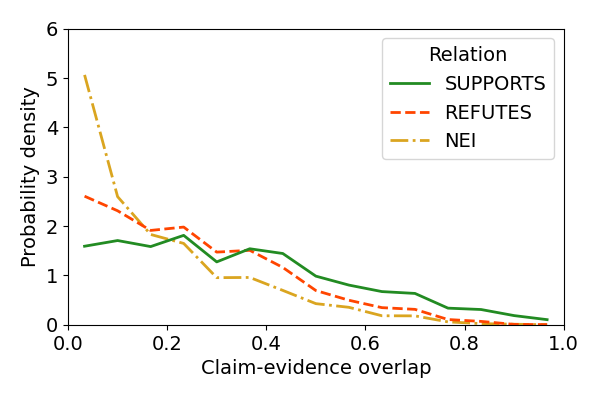} 
% \vspace*{-1.8\baselineskip}
\caption{\thedataset real}
\end{subfigure}
~
\begin{subfigure}{0.32\textwidth}
\includegraphics[width=1.05\linewidth]{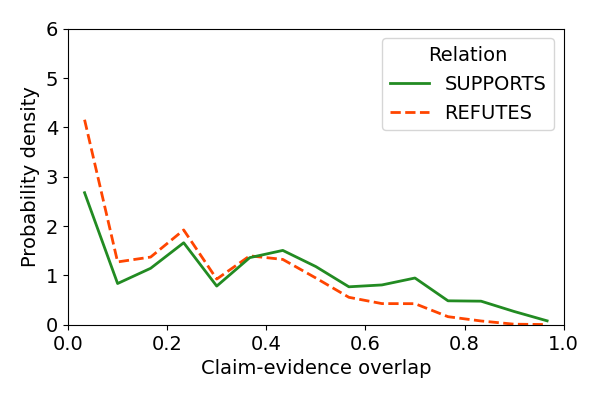}
% \vspace*{-1.8\baselineskip}
\caption{\thedataset synthetic}
\end{subfigure}
~
\begin{subfigure}{0.32\textwidth}
\includegraphics[width=1.05\linewidth]{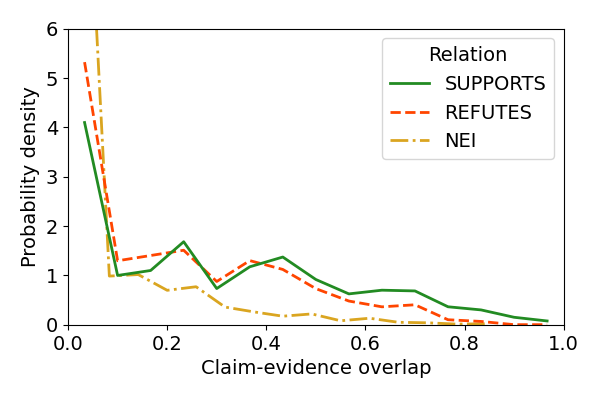} 
% \vspace*{-1.8\baselineskip}
\caption{FEVER}
\end{subfigure}

\small
% \vspace*{-.5\baselineskip}
\caption{Probability density function of claim-evidence overlap for different labels in the dataset. The overlap is computed as the ratio of mutual bigrams in the two sentences.}
%\vspace{-18pt}
\label{fig:overlap}
\end{figure*}

\subsection{Claim-only Classification}
Annotation artifacts are common in crowd-sourced sentence-pair inference datasets such as fact verification and NLI. 
Models can leverage these idiosyncrasies to achieve unexpectedly high performance when given only one sentence of the pair. For example, \citet{schuster-etal-2019-towards} showed that a claim-only classifier can obtain 61.7\% accuracy. The \thedataset dataset avoids this bias by pairing each claim with two contrastive contexts.

All claims in the \thedataset-synthetic are paired with one refuting and one supporting evidence, making it impossible for a claim-only to perform better than random. Each claim in the \thedataset-real is paired with one refuting or neutral evidence, in addition to a supporting one. To evaluate whether models can utilize lexical cues in claims, we train a claim-only classifier on \thedataset-real and find it to achieve 50\% accuracy---the same as always predicting $\SUP$.

\subsection{Claim-evidence Word Overlap}

Naturally, when pairing claims to evidence sentences, the overlapping words will be higher on average for claims with their supporting evidence. In \thedataset dataset, we want to minimize this bias in order to create challenging examples that require sentence-pair inference and cannot be solved by simple word matching techniques. Therefore, we asked annotators, when possible, to avoid copying exact phrases from the evidence to the claim (see \S\ref{sec:write_claims}). 

\figref{fig:overlap} shows the probability density function of bigram overlaps between the claim and evidence for each relation. Similar to FEVER, the overlap ratio of supporting pairs in the \thedataset dataset is only slightly higher than the one of refuting pairs. Also, the overlap ratio of the $\NEI$ pairs of the \thedataset real dataset is on average higher than FEVER.

% \subsection{Annotator Bias.}

%% file: appendix/setting.tex
\section{Experimental Setting}
We implement all our models with the HuggingFace Transformers library~\citep{Wolf2019HuggingFacesTS}. When comparing across training datasets of different sizes, we train the model for the same amount of update steps, upsampling the smaller datasets. We pick the checkpoint with the highest accuracy on the development set of the training task and report performance on the test set.
More details are available at \url{https://github.com/TalSchuster/VitaminC}

%% file: appendix/GPT.tex
\section{GPT-3 Evaluation}
\label{app:gpt}

The GPT-3 model has recently demonstrated impressive results in zero-shot and few-shot generation and classification tasks~\citep{brown2020language}. This 175B parameters language model was trained on billions of words from online sources, including the English Wikipedia. As result, it can be applied on many tasks without any further fine-tuning---instead, one need only provide a task-specific prefix (i.e., ``prompt'') with a few examples that direct the language model towards the desired output format. For example, GPT-3 achieves better than random results on ANLI with only a single example in the prompt, and over $40\%$ accuracy with 50 examples~\citep{brown2020language}.

We used OpenAI's beta API to query GPT-3. Due to our limited quota, we could not perform extensive experiments. Instead, we performed a qualitative evaluation using several examples from \thedataset test set for the claim extraction (factually consistent generation) and the fact verification tasks. Therefore, these results should be viewed as exploratory only.

\paragraph{GPT-3 for Claim Extraction.}
We examine a two-shot setting for the claim extraction task. The model is asked to convert a revision into a short claim that expresses the fact that is true after the edit. To guide the model for this task, we provide a prompt with two random examples from the \thedataset training set (see \figref{fig:gpt_claim_prompt}). One of the main concerns regarding large language models is the limited control it allows for ensuring that the facts in the generated output align with the source~\citep{schuster-etal-2020-limitations}. The generation tasks of \thedataset provide a useful test-bed for evaluating the factual consistency with the input. Importantly, our \thedataset-trained fact verification classifiers ($f_{\mathrm{verdict}}$) allow strong automatic evaluation for the factual agreement of the generation with the source.

We use GPT-3 to extract claims for four revisions with a sampling temperature value ($\mathcal{T}$) set to either $0$ or $0.7$. The zero value is recommended for maximizing the factual consistency as the model follows its most certain predictions. Using low temperature, however, can result in less fluent generations~\citep{Holtzman2020The}. Therefore, high values of $\mathcal{T}$ are also commonly used.

The results are reported in Tables \ref{tab:gen_claim_out} and \ref{tab:gen_claim_out_appendix}. With only two guiding examples, GPT-3 is able to follow the desired format and create a short claim. Yet, some of its generations follow $s_{t-1}$ instead of $s_{t}$ or add new, unsupported facts. $f_{\mathrm{verdict}}$ provides an indication for the factual correctness of the output. For example, it correctly classifies the output of the $\mathcal{T}=0.7$ setting for the top example in \tabref{tab:gen_claim_out} as ``Not Enough Information'' since GPT-3 reported about 20 deaths even though the input doesn't mention death numbers at all.

We expect GPT-3 to improve with longer prompts or fine-tuning and leave this to future research due to our limited quota. 

\paragraph{GPT-3 for Fact Verification.}
We also experiment with using GPT-3 few-shot classification capabilities for the fact verification task. We follow the ANLI few-shot format of \citet{brown2020language} and compose prompts with 6 examples (2 from each class) with random examples from \thedataset training set. We use only numerical examples to evaluate numerical claims (\figref{fig:gpt_fact_prompt_num}), and mixed examples for other claims (\figref{fig:gpt_fact_prompt}). We set $\mathcal{T}=0$ as recommended for classification.

\tabref{tab:fact_ver_gpt} summarizes the results. Even with only six examples, GPT-3 seems to perform significantly better than random. Yet, its verdict is wrong in several cases that can be easily classified by humans. For example, we find it to refrain from predicting a True/False verdict even when the evidence is clear. We observe this both for a date-based (line 3.2 in \tabref{tab:fact_ver_gpt}), numerical (lines 4.1-4.2), and entity-focused claims (line 5.2). 

To experiment with the sensitivity of the model to the provided context, we manually modified some of the examples to provide even stronger evidence. For example, while GPT-3's prediction for line 5.2 is acceptable as actually, Turner Broadcasting System merged with WarnerMedia in 1996, changing the evidence to another disconnected entity (The Walt Disney Company) did not change the prediction (line 5.3) as expected. Even when explicitly stating that there is no other owner GPT-3 didn't modify its verdict (line 5.4). Similarly, when evaluating the claim about the population of Beaverton being less than 90K, GPT-3 ignores the supporting evidence and outputs a false verdict (lines 1.4-1.5). Changing the claim to state ``approximately 86K'' instead of ``less than 90,000'' modified the prediction to ``Neither'' (line 1.6). Only repeating the exact same number as the evidence led to a true verdict (line 1.7).

\begin{figure*}[t]{ \tt \footnotesize \begin{tabularx}{\linewidth}{X} 
\toprule
Life Is Peachy: Life Is Peachy is the\{*| first -> second |\}studio album by the American nu metal band Korn, released on October 15, 1996 through both Immortal Records and Epic Records.\\
Claim: Life Is Peachy is Korn's second studio album.\\
\#\#\# \\
2020 coronavirus pandemic in Kerala: As of 14 March 2020, there are\{*| 19 -> 22 |\}confirmed cases of the virus and more than 4000 people are under surveillance in Kerala.\\
Claim: As of 14 March, there have been more than 20 confirmed COVID-19 cases in Kerala.\\
\#\#\# \\
<Visualized edit>\\
Calim: <prediction>\\
\bottomrule \end{tabularx}}

\caption{The prompt used for GPT-3 few-shot claim extraction.}
\label{fig:gpt_claim_prompt}

\end{figure*}

\begin{table*}[t]
\centering
\small
  \begin{tabular}{p{0.25cm}|p{4.6cm}|p{6cm}|p{0.9cm}|p{0.8cm}|p{0.8cm}}
\toprule
\# & Claim & Evidence & GPT-3 & Ours & Gold \\
\midrule
1.1 & \textbf{Less than 90,000} people live in Beaverton , Oregon & its population is estimated to be \textbf{91,757}, almost 14\% more than the 2000 census figure of 76,129 & False & False & False \\
\midrule
1.2 & \textbf{More than 90K} people live in Beaverton & its population is estimated to be \textbf{91,757}, almost 14\% more than the 2000 census figure of 76,129 & True & True & True \\
\midrule
1.3 & \textbf{More than 90K} people live in Beaverton & its population is estimated to be \textbf{86,205}, almost 14\% more than the 2000 census figure of 76,129 & \textcolor{red}{Neither} & False & False \\
\midrule
1.4 & \textbf{Less than 90,000} people live in Beaverton, Oregon & its population is estimated to be \textbf{86,205}, almost 14\% more than the 2000 census figure of 76,129 & \textcolor{red}{False} & True & True \\
\midrule
1.5 & \textbf{Less than 90,000} people live in Beaverton, Oregon & \textbf{Beaverton's} population is estimated to be \textbf{86,205} & \textcolor{red}{False} & True & True \\
\midrule
1.6 & \textbf{Approximately 86k} people live in Beaverto, Oregon & \textbf{Beaverton's} population is estimated to be \textbf{86,205} & \textcolor{red}{Neither} & True & True \\
\midrule
1.7 & \textbf{Approximately 86,205} people live in Beaverton, Oregon & \textbf{Beaverton's} population is estimated to be \textbf{86,205} & True & True & True \\
% \midrule
% 1.5 & \textbf{More than 90K} people live in Beaverton & \textbf{Beaverton's} population is estimated to be \textbf{86,205}, almost 14\% more than the 2000 census figure of 76,129 & \textcolor{red}{Neither} & False & False \\
%
\bottomrule \toprule
2.1 & Diego Corrales' father was Puerto Rican and his mother Dominican & Corrales was born to a African American father and a Mexican mother & False & False & False \\
\midrule
2.2 & Diego Corrales' father was Puerto Rican and his mother Dominican & Corrales was born to a Puerto Rican father and a Dominican mother & True & True & True \\
\bottomrule \toprule
3.1 & COVID-19 outbreak was identified \textbf{before December} & The outbreak was first identified in Wuhan, Hubei, China \textbf{in December} 2019 and recognized as a pandemic & False & False & False \\
\midrule
3.2 & COVID-19 outbreak was identified \textbf{before December} & The outbreak was first identified in Wuhan, Hubei, China \textbf{in 17 November} 2019 and recognized as a pandemic & \textcolor{red}{Neither} & True & True \\
\bottomrule \toprule
4.1 & There have been \textbf{more than 400} confirmed coronavirus cases in Germany & There have been \textbf{444} confirmed cases and 16 recoveries of coronavirus in Germany & \textcolor{red}{Neither} & True & True \\
 \midrule
4.2 & There have been \textbf{more than 400} confirmed coronavirus cases in Germany & There have been \textbf{less than 349} confirmed cases and 16 recoveries of coronavirus in Germany & \textcolor{red}{Neither} & False & False \\
\bottomrule \toprule
5.1 & Cartoon Network is owned by \textbf{Turner Broadcasting System} & Cartoon Network is an American pay television channel owned by \textbf{Turner Broadcasting System}, a subsidiary of AT\&T’s WarnerMedia & True & True & True \\
\midrule
5.2 & Cartoon Network is owned by \textbf{Turner Broadcasting System} & Cartoon Network is an American pay television channel owned by \textbf{Warner Bros.\ Entertainment}, a subsidiary of AT\&T’s WarnerMedia & \textcolor{red}{Neither} & False & False \\
\midrule
5.3 & Cartoon Network is owned by\textbf{ Turner Broadcasting System} & Cartoon Network is an American pay television channel owned by \textbf{The Walt Disney Company} & \textcolor{red}{Neither} & False & False \\
\midrule
5.4 & Cartoon Network is owned by \textbf{Turner Broadcasting System} & \textbf{The Walt Disney Company is the only owner} of Cartoon Network & \textcolor{red}{Neither} & False & False \\

\bottomrule
\end{tabular}
\caption{GPT-3 fact verification predictions on examples from the \thedataset test dataset (examples 1.5-1.7 and 5.3-5.4 were manually modified to examine the model's behavior). We follow the few-shot setting of \citet{brown2020language} for ANLI (see Figures~\ref{fig:gpt_fact_prompt} and \ref{fig:gpt_fact_prompt_num}). The bold spans are for presentation and are not part of the input. Our \thedataset-trained ALBERT classifiers predicted correctly on all these examples (though they weren't picked this way). The GPT-3 few-shot succeeds on some examples and even expresses sensitivity to evidence in lines 2.1-2.2. In several cases, however, GPT-3 abstains from a True/False verdict, even when provided with strong evidence (see ``Neither'' predictions). Line 1.4 shows an example where GPT-3's verdict is opposite of the provided evidence. Only when rephrasing the claim to exactly overlap with the evidence, it predicts an agreement.} \label{tab:fact_ver_gpt}
\end{table*}

\begin{figure*}[t]{ \tt \footnotesize \begin{tabularx}{\linewidth}{X} 
\toprule
Manchester is a major city and metropolitan borough in Greater Manchester, England, with a population of 545,500 as of 2017 (5th most populous English district). \\
Question: Manchester had a population of more than 540,000 in 2017 and was the 5th most populous English district. True, False, or Neither? True \\
\#\#\# \\
As of March 2018, the apps have achieved more than 8 billion downloads.\\
Question: Talking Tom and Friends apps have less than 8 billion downloads. True, False, or Neither? False\\
\#\#\# \\
He won the Premier League in 2018.\\
John Stones won both the Premier League and EFL Cup in 2018. True, False, or Neither? Neither\\
\#\#\# \\
Neck Deep are a emo band.\\
Question: Neck deep is an emo band. True, False, or Neither? True\\
\#\#\# \\
Critics generally gave The Final Frontier mixed to poor reviews.\\
Question: The film Star Trek V: The Final Frontier got negative reviews only. True, False, or Neither? False\\
\#\#\# \\
The series was favorably compared to the HBO series The Jinx and the podcast Serial.\\
Question: The follow-up of the series Making a Murderer, was released in 2018. True, False, or Neither? Neither\\
\#\#\# \\
<Examined evidence>\\
Question: <Examined claim>. True, False, or Neither? <prediction>\\
\bottomrule \end{tabularx}}

\caption{The prompt used for GPT-3 few-shot fact verification predictions on \textbf{non-numerical} claims (examples 2 and 4 in \tabref{tab:fact_ver_gpt}. We follow the few-shot setting of \citet{brown2020language} for ANLI.}
\label{fig:gpt_fact_prompt}

\end{figure*}

\begin{figure*}[t]{ \tt \footnotesize \begin{tabularx}{\linewidth}{X} 
\toprule
Manchester is a major city and metropolitan borough in Greater Manchester, England, with a population of 545,500 as of 2017 (5th most populous English district).\\
Question: Manchester had a population of more than 540,000 in 2017 and was the 5th most populous English district. True, False, or Neither? True\\
\#\#\# \\
As of March 2018, the apps have achieved more than 8 billion downloads.\\
Question: Talking Tom and Friends apps have less than 8 billion downloads. True, False, or Neither? False\\
\#\#\# \\
As of January 2015, JFC had a total of more than 3,000 stores worldwide, with system-wide retail sales totaling 82.1 billion pesos for the fiscal year 2011.\\
Question: Jollibee had a total of more than 20,000 stores worldwide after January 2016. True, False, or Neither? Neither\\
\#\#\# \\
As of March 2018, the apps have achieved more than 8 billiobn downloads.\\
Question: Talking Tom and Friends apps have over 8 billion downloads. True, False, or Neither? True\\
\#\#\# \\
Bet365 has more than 35 million customers globally.\\
Question: Bet365 has less than 30 million customers worldwide. True, False, or Neither? False\\
\#\#\# \\
The series was favorably compared to the HBO series The Jinx and the podcast Serial.\\
Question: The follow-up of the series Making a Murderer, was released in 2018. True, False, or Neither? Neither\\
\#\#\# \\
<Examined evidence>\\
Question: <Examined claim>. True, False, or Neither? <prediction>\\
\bottomrule \end{tabularx}}

\caption{The prompt used for GPT-3 few-shot fact verification predictions on \textbf{numerical} claims (examples 1 and 3 in \tabref{tab:fact_ver_gpt}.}
\label{fig:gpt_fact_prompt_num}

\end{figure*}

%% file: appendix/more_res.tex
\section{Complementary Experiments}

We report fact verification results with a fine-tuned BERT-base~\citep{devlin-etal-2019-bert} model in \tabref{tab:res_fact_ver_bert}. We find ALBERT-base to outperform BERT-base on most of the evaluated datasets. ALBERT-xlarge performed better than the two base models in all datasets except for Triggers. The Triggers dataset is very small (186 examples) and contains some unnaturally looking claims, which could explain the high variance across models.

\input{appendix/tabels/bert_res}

%% file: appendix/tabels/bert_res.tex
\begin{table*}[!h]
\centering
\resizebox{2\columnwidth}{!}{%
\begin{tabular}{l|l|cccc|cccc|c}
\toprule
Model & Train dataset & VitC real & VitC syn & FEVER & MNLI &  Adversarial & Symmetric  & Triggers  & ANLI & Contrast \\

\midrule

\multirow{6}{*}{BERT-base}

& FEVER                &  60.55 & 71.35 &	87.16 &	61.90 &	52.09 &	73.60 &	69.89  &	\textbf{34.53}  & 54.05       \\
& MNLI &           46.31 &	69.01 &	70.06 &	83.80 &	50.13 &	73.88 &	65.05  &	26.88 & 51.92 \\
& FEVER + MNLI &   56.24 &	81.80 &	\textbf{95.59} &	\textbf{85.06} &	63.05 &	85.11 &	37.63 &	29.63 & 60.63\\
\cmidrule{2-11}
& VitC                 &  \textbf{85.80} & 90.63 &	74.21 &	66.66 &	\textbf{76.24} &	90.17 &	63.98 &	33.19 & 72.49         \\
& VitC + MNLI &    84.47 &	\textbf{91.00} &	74.88 &	83.70 &	63.05 &	84.55 &	66.13 &	31.00 & 84.88\\
& VitC + FEVER         &  84.72 & 89.16 &	87.55 &	69.28 &	64.75 &	\textbf{90.73} &	\textbf{72.58}  &	34.06 & 84.01           \\

\bottomrule

\end{tabular}
}%
\caption{Fact verifcation Complementrary results for \tabref{tab:res_fact_ver} with a BERT-base model.}  \label{tab:res_fact_ver_bert}
    % \vspace{-10pt}

\end{table*}

%% file: appendix/outputs.tex
\section{Example Outputs}\label{app:examples}

We provide examples of predicted word-level rationales in \tabref{tab:output_rationale} and of outputs for the two generation tasks in Tables \ref{tab:gen_rev_out} and \ref{tab:gen_claim_out_appendix}.

% \subsection{Word-level Rationales}
\newcommand{\mask}[1]{\addbox{#1}}
\begin{table*}[!h]
\centering
\small
\begin{tabular}{l|l}
\toprule
Claim    & the youtube channel chuchu tv is placed 42nd and has more than 25 million subscribers .                                                \\
Evidence & chuchu tv is the \mask{43rd} most \mask{subscribed youtube} channel in the world , with over \mask{20 million subscribers} .           \\ \midrule
Claim    & the ramus has sold less than 4.5 million albums worldwide .                                                                            \\
Evidence & the rasmus has \mask{sold 5 million albums worldwide} , 310,000 copies in their native finland alone .                                 \\ \midrule
Claim    & darren randolph is spanish .                                                                                                           \\
Evidence & humes dated \mask{irish} footballer darren randolph in 2005 .                                                                          \\ \midrule
Claim    & astravyets is near vilnius .                                                                                                           \\
Evidence & his father may have migrated to the us in the 1860s from \mask{astravyets near vilnius} .                                              \\ \midrule
Claim    & the pace controlling stamina meter is a new feature in the game series .                                                               \\
Evidence & \mask{new} to the \mask{series} is a three-tier \mask{stamina meter} which \mask{controls} the \mask{pace} of a match .                \\ \midrule
Claim    & the movie will be released on 25 november 2015 .                                                                                       \\
Evidence & [...] are producing the film which columbia pictures will \mask{release on november 25} , \mask{2015}. \\ \bottomrule
\end{tabular}
\caption{Example masks produced by the word-level rationale model for identifying \mask{anchoring words} in the evidence that are responsible for the classifiers verdict regarding the claim. Masking these words leads the classifier to predict $\NEI$ instead of what would have been $\SUP$ or $\REF$.} \label{tab:output_rationale}
    % \vspace{-16pt}
\end{table*}

% \subsection{Factually Consistent Generation}

%\include{appendix/tabels/gen_examples}

\begin{table*}[t]
\centering
\small
  \begin{tabular}{p{1cm}|p{13cm}}
\toprule
Claim    &  Stephen Bruner was born on October 19 , 1984 .   \\
\cmidrule{2-2} 
$s_{t-1}$ & Stephen Bruner ( born October 18 , 1984 ) , better known by his stage name Thundercat , is an American multi-genre bass player , producer and singer fr     om Los Angeles , California .\\
\cmidrule{2-2} 
Output & Stephen Bruner ( born October 19, 1984 ), better known by his stage name Thundercat, is an American multi-genre bass player, producer and singer from Los Angeles, California.\\
\cmidrule{2-2} 
$s_t$ & Stephen Bruner ( born October 19 , 1984 ) , better known by his stage name Thundercat , is an American multi-genre bass player , producer and singer from Los Angeles , California .\\
\midrule

Claim    &   The House that Jack Built has an approval rate of over 45 \% and over 13 critics .  \\
\cmidrule{2-2} 
$s_{t-1}$ & On Rotten Tomatoes , the film has an approval rating of 42 \% , based on 12 critics , and an average rating of 5.9/10 \\
\cmidrule{2-2} 
Output & On Rotten Tomatoes, the film has an approval rating of 47 \%, based on 14 critics, and an average rating of 5.9/10.\\
\cmidrule{2-2} 
$s_t$ & On Rotten Tomatoes , the film has an approval rating of 47 \% , based on 15 critics , and an average rating of 6.3/10. \\
\midrule

Claim    &  Cartoon Network is owned by Warner Bros. Entertainment .  \\
\cmidrule{2-2} 
$s_{t-1}$ & Cartoon Network is an American pay television channel owned by Turner Broadcasting System , a subsidiary of AT \& T 's WarnerMedia . \\
\cmidrule{2-2} 
Output & Cartoon Network is an American pay television channel owned by Warner Bros. Entertainment, a subsidiary of AT \& T's WarnerMedia. \\
\cmidrule{2-2} 
$s_t$ & Cartoon Network is an American pay television channel owned by Warner Bros. Entertainment , a subsidiary of AT \& T 's WarnerMedia . \\

\midrule

Claim    &  The New York Times reviewed the film Mad Max : Fury Road .  \\
\cmidrule{2-2} 
$s_{t-1}$ & Lindsay Bahr of The Associated Press wrote , `` Miller has reminded us that blockbusters have the potential to not only be art , but radically visionary – even the fourth in a series . \\
\cmidrule{2-2} 
Output & Lindsay Bahr of The New York Times wrote, `` Miller has reminded us that blockbusters have the potential to not
 only be art, but radically visionary – even the fourth in a series. \\
\cmidrule{2-2} 
$s_t$ & The New York Times wrote , `` Miller has reminded us that blockbusters have the potential to not only be art , but radically visionary – even the fourth in a series . \\

\bottomrule
\end{tabular}
\caption{Example outputs of the BART-base used for generating factually consistent revisions given the old version $s_{t-1}$ and the updated claim we wish to support.  The ``ground-truth'' $s_{t}$ is provided for reference. } \label{tab:gen_rev_out}
\end{table*}

%%%% claim gen

\begin{table*}[!h]
\centering
\small
    \resizebox{2.05\columnwidth}{!}{%
%   \begin{tabular}{p{1.8cm}|p{13cm}}
  \begin{tabular}{p{1.8cm}|p{12.5cm}|p{0.5cm}}
% \multicolumn{3}{r}{$f_{\mathrm{verdict}}(c,s_t)$}  \\
\toprule

$\revtup$    & \multicolumn{2}{l}{WWE 2K15: As of August 2015 , WWE 2K15 has shipped over\{*| six -> 7.5 |\}million units.}  \\
\cmidrule{2-3} 
BART (VitC) & \multicolumn{1}{l}{WWE 2K15 sold more than 7 million units. \qquad \qquad \qquad \qquad \qquad \qquad \qquad \quad $\vert$ \  $f_{\mathrm{verdict}}(c,s_t)=$}& $\SUP$\\
\cmidrule{2-2} 
GPT-3 $\mathcal{T}$=$0$ & As of August 2015, WWE 2K15 has shipped over seven million units & $\SUP$\\
\cmidrule{2-2} 
GPT-3 $\mathcal{T}$=$0.7$ & As of August 2015, WWE 2K15 has shipped over seven million units & $\SUP$\\
\cmidrule{2-2} 
Reference & \textcolor{darkgreen}{By August 2015 , WWE 2K15 shipped over 7.5 million units .}& $\SUP$\\

\midrule
% $\revtup$    & \multicolumn{2}{l}{\multirowcell{2}[0ex][l]{Diego Corrales: Corrales was born to a\{*| \textbf{Puerto Rican -> African American} |\}father and a\\\{*| \textbf{Dominican -> Mexican} |\}mother .}}\\
$\revtup$    & \multicolumn{2}{l}{\multirowcell{2}[0ex][l]{Pat Jennings: He has played for League of Ireland clubs UCDDerry and is now at\\ \{*| Shamrock Rovers -> Dublin |\}.}}\\
& \multicolumn{2}{c}{}\\
\cmidrule{2-3} 
BART (VitC) & Pat Jennings is currently playing for Dublin club UCDDerry. & $\SUP$\\
\cmidrule{2-2} 
GPT-3 $\mathcal{T}$=$0$ & He has played for League of Ireland clubs UCD and is now at Shamrock Rovers & $\REF$\\
\cmidrule{2-2} 
GPT-3 $\mathcal{T}$=$0.7$ & He played for Shamrock Rovers and is now at Dublin & $\REF$\\
\cmidrule{2-2} 
Reference & \textcolor{darkgreen}{Pat Jennings currently plays for the Dublin club .}& $\SUP$\\

\bottomrule
\end{tabular}
}%
\caption{Additional examples for \tabref{tab:gen_claim_out}. Example outputs for extracting claims that express the factual change in a Wikipedia revision. The BART-base model is trained on \thedataset data and GPT-3 is applied in a 2-shot setting with a temperature of 0 or 0.7. The revision $\revtup$ is given to the model as a single sentence visualization where the edits are between curly brackets, preceded by the article's title.  The human-written claim is provided for reference. The prediction of our ALBERT-xlarge \thedataset-trained model $f_{\mathrm{verdict}}(c,s_t)$ on the generated claim against $s_t$ is also reported in the rightmost column.} \label{tab:gen_claim_out_appendix}
\end{table*}